\pdfoutput=1

\documentclass[twocolumn]{article}
\usepackage[numbers,square,sort&compress]{natbib}
\usepackage[export]{adjustbox}
\usepackage{algorithm}
\usepackage{algpseudocode}
\usepackage{amsmath}
\usepackage{amssymb}
\usepackage{array}
\usepackage{authblk}
\usepackage{dcolumn}
\usepackage{enumitem}
\usepackage{fix-cm}
\usepackage[spacem,multidot]{grffile}
\usepackage{listings}
\usepackage{multirow}
\usepackage{pgffor}
\usepackage{placeins}
\usepackage[labelformat=empty]{subfig}
\usepackage{tabularx}
\usepackage{multirow}
\usepackage{tikz}
\usepackage{times}
\usepackage{xcolor}
\usepackage{xspace}
\usepackage{graphicx}
\usepackage[
pagebackref=true,
colorlinks,
bookmarks=false]{hyperref} 

\graphicspath{{./figures-flattened/}, {./original_images/}, {./imp_figures/}}
\newcommand{\real}{\mathbb{R}}

\newcommand{\be}{\mathbf{e}}

\newcommand{\bg}{\mathbf{g}}

\newcommand{\bu}{\mathbf{u}}

\newcommand{\bw}{\mathbf{w}}
\newcommand{\bx}{\mathbf{x}}

\newcommand{\TV}{{TV^\beta}}

\makeatletter
\DeclareRobustCommand\onedot{\futurelet\@let@token\@onedot}
\def\@onedot{\ifx\@let@token.\else.\null\fi\xspace}
\newcommand{\eg}{\emph{e.g}\onedot}

\newcommand{\etal}{\emph{et al}\onedot}

\def\sectionautorefname~#1\null{%
  Sect.~#1\null
}
\def\subsectionautorefname~#1\null{%
  Sect.~#1\null
}
\def\subsubsectionautorefname~#1\null{%
  Sect.~#1\null
}
\def\equationautorefname~#1\null{%
  Eq.~(#1)\null
}
\def\figureautorefname~#1\null{%
  Fig.~#1\null
}

\tikzset{
 image label/.style={
   fill=white,
   text=black,
   font=\footnotesize,
   anchor=south east,
   xshift=-0.1cm,
   yshift=0.1cm,
   at={(0,0)}
 }
}

\newcommand{\puti}[2]
{%
\begin{tikzpicture}
\node[anchor=south east,inner sep=0] at (0,0) {#2};
\node[image label]{#1};
\end{tikzpicture}%
}
\newlength{\x}
\newlength{\y}

\newcommand{\includegrid}[5][0pt 0pt 0pt 0pt]{%
\setlength{\x}{#2}%
\setlength{\y}{\dimexpr\x/#3\relax}%
\begin{tikzpicture}%
\clip (0,0) rectangle (\x,\x);
\node (fig1) at (.5\x,.5\x) {%
\adjincludegraphics[width=\x,interpolate=false,trim={#1},clip]{{#5}}%
};
\draw[step=\y,green,thick] (0,0) grid (\x,\x);
\node[fill=white,anchor=south west] at (2pt,2pt) {#4};
\end{tikzpicture}%
}

\newcommand{\includelabel}[3]{%
\setlength{\x}{#1}%
\begin{tikzpicture}%
\clip (0,0) rectangle (\x,\x);
\node (fig1) at (.5\x,.5\x) {\includegraphics[width=\x]{{#3}}};
\node[fill=white,anchor=south west] at (2pt,2pt) {#2};
\end{tikzpicture}%
}

\usepackage[margin=1in]{geometry}

\begin{document}

\title{Visualizing deep convolutional neural networks using natural pre-images}
\author{Aravindh Mahendran\thanks{aravindh@robots.ox.ac.uk}}
\author{Andrea Vedaldi\thanks{vedaldi@robots.ox.ac.uk}}
\affil{University of Oxford}
\maketitle

\begin{abstract}
Image representations, from SIFT and bag of visual words to Convolutional Neural Networks (CNNs) are a crucial component of almost all computer vision systems. However, our understanding of them remains limited. In this paper we study several landmark representations, both shallow and deep, by a number of complementary visualization techniques. These visualizations are based on the concept of ``natural pre-image'', namely a natural-looking image whose representation has some notable property. We study in particular three such visualizations: inversion, in which the aim is to reconstruct an image from its representation, activation maximization, in which we search for patterns that maximally stimulate a representation component, and caricaturization, in which the visual patterns that a representation detects in an image are exaggerated. We pose these as a regularized energy-minimization framework and demonstrate its generality and effectiveness. In particular,  we show that this method can invert representations such as HOG more accurately than recent alternatives while being applicable to CNNs too.  Among our findings, we show that several layers in CNNs retain photographically accurate information about the image, with different degrees of geometric and photometric invariance.
\end{abstract}

\section{Introduction}\label{s:intro}

\begin{figure}[t]\centering%
\begin{tikzpicture}
\clip (0,0) rectangle (1\columnwidth,1\columnwidth);
\node[anchor=south west,inner sep=0] at (0,0) {
\includegraphics[width=1\columnwidth]{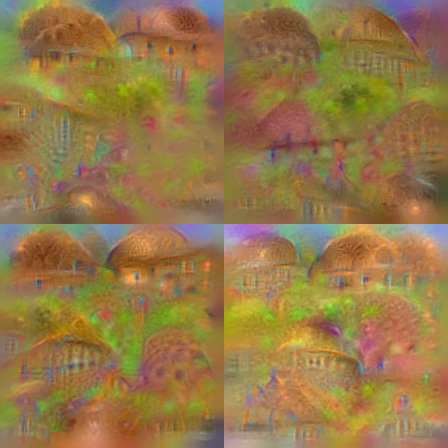}
};
\node[anchor=south east,inner sep=0] at (.8\columnwidth-5pt,0+5pt) {
\includegraphics[width=0.25\columnwidth]{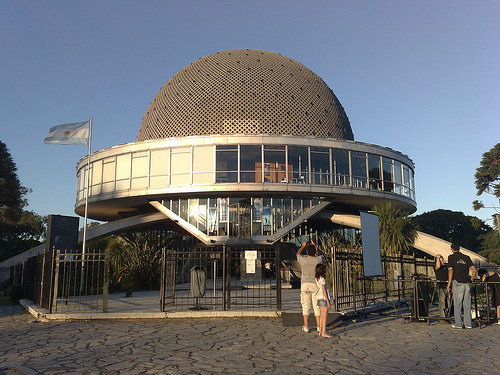}
};
\end{tikzpicture}
\caption{Four reconstructions of the bottom-right image obtained from the 1,000 code extracted from the last fully connected layer of the VGG-M CNN~\cite{chatfield14return}. This figure is best viewed in color.}\label{f:splash}
\end{figure}

Most image understanding and computer vision methods do not operate directly on images, but on suitable image representations. Notable examples of representations include  textons~\cite{leung01representing}, histogram of oriented gradients (SIFT~\cite{lowe04distinctive} and HOG~\cite{dalal05histograms}), bag of visual words~\cite{csurka04visual}\cite{sivic03video}, sparse~\cite{yang10supervised} and local coding~\cite{wang10locality-constrained}, super vector coding~\cite{zhou10image}, VLAD~\cite{jegou10aggregating}, Fisher Vectors~\cite{perronnin06fisher}, and, lately, deep neural networks, particularly of the convolutional variety~\cite{krizhevsky12imagenet,zeiler14visualizing,sermanet14overfeat:}.
While the performance of representations has been improving significantly in the past few years, their design remains eminently empirical. This is true for shallower hand-crafted features such as HOG or SIFT and even more so for the latest generation of deep representations, such as deep Convolutional Neural Networks (CNNs), where millions of parameters are learned from data.  A consequence of this complexity is that our understanding of such representations is limited.

In this paper, with the aim of obtaining a better understanding of representations, we develop a family of methods to investigate CNNs and other image features by means of visualizations. All these methods are based on the common idea of seeking natural-looking images whose representations are notable in some useful sense. We call these constructions \emph{natural pre-images} and propose a unified formulation and algorithm to compute them (\autoref{s:method}). 

Within this framework, we explore three particular types of visualizations. In the first type, called \textbf{inversion} (\autoref{s:results-inversion}), we compute the ``inverse'' of a representation (\autoref{f:splash}). We do so by modelling a representation as a function $\Phi_0 = \Phi(\bx_0)$ of the image $\bx_0$. Then, we attempt to recover the image from the information contained only in the code $\Phi_0$. Notably, most representations $\Phi$ are \emph{not} invertible functions; for example, a representation that is \emph{invariant} to nuisance factors such as viewpoint and illumination removes this information from the image. Our aim is to characterize this loss of information by studying the equivalence class of images $\bx^*$ that share the same representation $\Phi(\bx^*) = \Phi_0$. 

In \textbf{activation maximization}  (\autoref{s:results-activation-maximization}), the second visualization type, we look for an image $\bx^*$ that maximally excites a certain component $[\Phi(\bx)]_i$ of the representation. The resulting image is representative of the visual stimuli that are selected by   that component and helps understand its ``meaning'' or function. This type of visualization is sometimes referred to as ``deep dream'' as it can be interpreted as the result of the representation ``imagining'' a concept.

In our third and last visualization type, which we refer to as \textbf{caricaturization} (\autoref{s:results-activation-highlighting}), we modify an initial image $\bx_0$ to exaggerate any pattern that excites the representation $\Phi(\bx_0)$.  Differently from activation maximization, this visualization method emphasizes the meaning of combinations of representation components that are active together.

Several of these ideas have been explored by us and others in prior work as detailed in \autoref{s:related}. In particular, the idea of visualizing representations using pre-images has been investigated in connection with neural networks since at least the work of Linden~\etal~\cite{linden89inversion}.

Our first contribution is to introduce the idea of a \emph{natural pre-image}~\cite{mahendran15understanding}, i.e. to restrict reconstructions to the set of natural images. While this is difficult to achieve in practice, we explore different regularization methods (\autoref{s:regularization}) that can work as a proxy, including regularizers using the Total Variation (TV) norm of the image. We also explore an indirect regularization method, namely the application of random jitter to the reconstruction as suggested by Mordvintsev~\etal~\cite{mordvintsev15inceptionism:}.

Our second contribution is to consolidate different visualization and representation types, including inversion, activation maximization, and caricaturization, in a common framework (\autoref{s:method}). We propose a single algorithm applicable to a large variety of representations, from SIFT to very deep CNNs, using essentially a single set of parameters. The algorithm is based on optimizing an energy function using gradient descent and back-propagation through the representation architecture. 

Our third contribution is to apply the three visualization types to the study of several different representations. First, we show that, despite its simplicity and generality, our method recovers significantly better reconstructions for shallow representations such as HOG compared to recent alternatives~\cite{vondrick13hoggles:} (\autoref{s:results-inversion-classical}). In order to do so, we also rebuild the HOG and DSIFT representations as equivalent CNNs, simplifying the computation of their derivatives as required by our algorithm (\autoref{s:shallow-networks}). Second, we apply inversion (\autoref{s:result-inversion-cnn}), activation maximization (\autoref{s:results-activation-maximization}), and caricaturization (\autoref{s:results-activation-highlighting}) to the study of CNNs, treating each layer of a CNN as a different representation, and studying different state-of-the-art architectures, namely AlexNet, VGG-M, and VGG very deep (\autoref{s:deep-networks}). As we do so, we emphasize a number of general properties of such representations, as well as differences between them. In particular, we study the effect of depth on representations, showing that CNNs gradually build increasing levels of invariance and complexity, layer after layer.

Our findings are summarized in \autoref{s:summary}. The code for the experiments in this paper and extended visualizations are available at~\url{http://www.robots.ox.ac.uk/~vgg/research/invrep/index.html}. This code uses the open-source MatConvNet toolbox~\cite{matconvnet2014vedaldi} and publicly available copies of the models to allow for easy reproduction of the results.

This paper is a substantially extended version of~\cite{mahendran15understanding}, which introduced the idea of natural pre-image, but was limited to visualization by inversion.

\section{Related work}\label{s:related}

With the development of modern visual representations, there has been an increasing interest in developing visualization methods to understand them. Most of the recent contributions~\cite{mordvintsev15inceptionism:,yosinksi15understanding} build on the idea of \emph{natural} pre-images introduced in~\cite{mahendran15understanding}, extending or applying it in different ways. In turn, this work is based on several prior contributions that have used pre-images to understand neural networks and classical computer vision representations such as HOG and SIFT. The rest of the section discusses these relationships in detail.

\subsection{Natural pre-images}

Mahendran~\etal~\cite{mahendran15understanding} note that not all pre-images are equally interesting in visualization; instead, more meaningful results can be obtained by restricting pre-images to the set of natural images. This is particularly true in the study of discriminative models such as CNNs that are essentially ``unspecified'' outside the domain of natural images used to train them. While capturing the concept of natural images in an algorithm is difficult in practice, Mahendran~\etal proposed to use simple natural image priors as a proxy. They formulated this approach in a regularized energy minimization framework. Among these, the most important regularizer was the quadratic norm\footnote{It is referred to as TV norm in~\cite{mahendran15understanding} but for $\beta=2$ this is actually the quadratic norm.} of the reconstructed image (\autoref{s:regularization}).

The visual quality of pre-images can be further improved by introducing complementary regularization methods. Google's ``inceptionism''~\cite{mordvintsev15inceptionism:}, for example, contributed the idea of regularization through jittering: they shift the pre-image  randomly during optimization, resulting in sharper and more vivid reconstructions. The work of Yosinksi~\etal~\cite{yosinksi15understanding} used yet another regularizer: they applied Gaussian blurring and clipped pixels that have small values or that have a small effect on activating components in a CNN representation to zero.

\subsection{Methods for finding pre-images}\label{s:preimage-methods}

The use of pre-images to visualize representations has a long history.  Simonyan \etal~\cite{simonyan14deep} applied this idea to recent CNNs and optimized, starting from random noise and by means of back-propagation and gradient descent, the response of individual filters in the last layer of a deep convolutional neural network -- an example of activation maximization. Related energy-minimization frameworks were adopted by~\cite{mahendran15understanding,mordvintsev15inceptionism:,yosinksi15understanding} to visualize recent CNNs. Prior to that, very similar methods were applied to early neural networks in~\cite{williams86inverting,linden89inversion,lee94inverse,lu99inverting}, using gradient descent or optimization strategies based on sampling.

Several pre-image methods alternative to energy minimization have been explored as well. Nguyen~\etal~\cite{nguyen15deep} used genetic programming to generate images that maximize the response of selected neurons in the very last layer of a modern CNN, corresponding to an image classifier.  Vondrick~\etal~\cite{vondrick13hoggles:} learned a regressor that, given a HOG-encoded image patch, reconstructs the input image. Weinzaepfel~\etal~\cite{weinzaepfel11reconstructing} reconstructed an image from SIFT features using a large vocabulary of patches to invert individual detections and blended the results using Laplace (harmonic) interpolation. Earlier works~\cite{jensen99inversion,varkonyi-koczy05observer} focussed on inverting networks in the context of dynamical systems and will not be discussed further here.

The DeConvNet method of Zeiler and Fergus~\cite{zeiler14visualizing} ``transposes'' CNNs to find which image patches are responsible for certain neural activations.  While this transposition operation applied to CNNs is somewhat heuristic, Simonyan~\etal~\cite{simonyan14deep} suggested that it approximates the derivative of the CNN and that, thereby, DeConvNet is analogous to one step of the backpropagation algorithm used in their energy minimization framework. A significant difference from our work is that in DeConvNet the authors transfer the pattern of activations of max-pooling layers from the direct CNN evaluation to the transposed one, therefore copying rather than inferring this geometric information during reconstruction. 

A related line of work~\cite{dosovitskiy15inverting,bishop95neural} is to learn a second neural network to act as the inverse of the original one. This is difficult because the inverse is usually not unique. Therefore, these methods may regress an ``average pre-image'' conditioned on the target representation, which may not be as effective as sampling the pre-image if the goal is to characterize representation ambiguities. One advantage of these methods is that they can be significantly faster than energy minimization.

Finally, the vast family of auto-encoder architectures~\cite{hinton06reducing} train networks together with their inverses as a form of auto-supervision; here we are interested instead in visualizing feed-forward and discriminatively-trained CNNs now popular in computer vision.

\subsection{Types of visualizations using pre-images}\label{s:visualization-types}

Pre-images can be used to generate a large variety of complementary visualizations, many of which have been applied to a variety of representations.

The idea of \textbf{inverting representations} in order to recover an image from its encoding was used to study SIFT in the work of~\cite{weinzaepfel11reconstructing}, Local Binary Descriptors by d'Angelo~\etal~\cite{dalengo2012beyond}, HOG in~\cite{vondrick13hoggles:} and bag of visual words descriptors in Kato~\etal~\cite{Kato_2014_CVPR}. \cite{mahendran15understanding} looked at the inversion problem for HOG, SIFT, and recent CNNs; our method differs significantly from the ones above as it addresses many different representations using the same energy minimization framework and optimization algorithm. In comparison to existing inversion techniques for dense shallow representations such as HOG~\cite{vondrick13hoggles:}, it is also shown to achieve superior results, both quantitatively and qualitatively.

Perhaps the first to apply \textbf{activation maximization} to recent CNNs such as AlexNet~\cite{krizhevsky12imagenet} was the work of Simonyan~\etal~\cite{simonyan14deep}, where this technique was used to maximize the response of neural activations in the last layer of a deep CNN. Since these responses are learned to correspond to specific object classes, this produces versions of the object as conceptualized by the CNN,  sometimes called ``deep dreams''.  Recently,~\cite{mordvintsev15inceptionism:} has  generated similar visualizations for their inception network and Yosinksi~\etal~\cite{yosinksi15understanding} have applied activation maximization to visualize not only the last layers of a CNN, but also intermediate representation components. Related extensive component-specific visualizations were conducted in~\cite{zeiler14visualizing}, albeit in their DeConvNet framework. The idea dates back to at least~\cite{erhan09visualizing}, which introduced activation maximization to visualize deep networks learned from the MNIST digit dataset.

The first version of \textbf{caricaturization} was explored in~\cite{simonyan14deep} to maximize image features corresponding to a particular object class, although this was ultimately used to generate saliency maps rather than to generate an image. The authors of~\cite{mordvintsev15inceptionism:} extensively explored caricaturization in their ``inceptionism'' research with two remarkable results. The first was to show which visual structures are captured at different levels in a deep CNN. The second was to show that CNNs can be used to generate aesthetically pleasing images.

In addition to these three broad visualization categories, there are several others which are more specific. In DeConvNet~\cite{zeiler14visualizing} for example, visualizations are obtained by activation maximization. They search in a large dataset for an image that causes a given representation component to activate maximally. The network is then evaluated feed-forward and the location of the max-pooling activations is recorded. Combined with the transposed ``deconvolutional network'', this information is used to generate crisp visualizations of the excited neural paths. However, this differs from both inversion and activation maximization in that it uses of information beyond that contained in the representation output itself.

\subsection{Activation statistics}\label{s:statmatch}

In addition to inversion, activation maximization, and caricaturization, the pre-image method can be extended to several other visualization types with either the goal of understanding representations or of generating images for other purposes. Next, we discuss a few notable cases.

First, representations can be used as \emph{statistics that describe a class of images}. This idea is rooted in the seminal work of Julesz~\cite{julesz81textons} that used the statistics of simple filters to describe \textbf{visual textures}. Julesz' ideas were framed probabilistically by Zhu and Mumford~\cite{zhu98filters} and their generation-by-sampling framework was later approximated by Portilla and Simoncelli~\cite{portilla00aparametric} as a pre-image problem which can be seen as a special case of the inversion method discussed here. More recently, Gatys~\etal~\cite{gatys15texture} showed that the results of Portilla and Simoncelli can be dramatically improved by replacing their wavelet-based statistics with the empirical correlation between deep feature channels in the convolutional layers of CNNs.

Gatys~\etal further extended their work in \cite{gatys15aneural} with the idea of \textbf{style transfer}. Here a pre-image is found that simultaneously (1) reproduces the deep features of a reference ``content image'' (just like the inversion technique explored here) while at the same time (2) reproducing the correlation statistics of shallower features of a second reference ``style image'', treated as a source of texture information. This can be interpreted naturally in the framework discussed here as visualization by inversion where the natural image prior is implemented by ``copying'' the style of a visual texture. In comparison to the approach here, this generally results in more pleasing images. For understanding representations, such a technique can be used to encourage the generation of very different images that share a common deep feature representation, and that therefore may reveal interesting invariance properties of the representation.

Finally, another difference between the work of Gatys~\etal~\cite{gatys15texture,gatys15aneural} and the analysis in this paper is that they transfer information from several layers of the CNN simultaneously, whereas here we focus on individual layers, or even single feature components. Thus the two approaches are complementary. In their case, there is no need to add an explicit natural image prior as we do as this information is incorporated in the low-level CNN statistics that they import in style/texture transfer. As shown in the experiments, a naturalness prior is however important when the goal is to visualize deep features without biasing the reconstruction using this shallower information at the same time.

\subsection{Fooling representations}\label{s:fooling}

A line of research related to visualization by pre-images is that of ``fooling representations''. Here the goal is to generate images that a representation assigns to a particular category despite having distinctly incompatible semantics. Some of these methods look for \emph{adversarial perturbations} of a source image. For instance, Tatu~\etal~\cite{tatu11exploring} show that it is possible to make any two images look nearly identical in SIFT space up to the injection of  adversarial noise in the data. The complementary effect was demonstrated for CNNs by Szegedy~\etal~\cite{szegedy13intriguing}, where an imperceptible amount of adversarial noise was shown to change the predicted class of an image to any desired class. The latter observation was confirmed and extended by~\cite{nguyen15deep}. The instability of representations appear in contradiction with results in~\cite{weinzaepfel11reconstructing,vondrick13hoggles:,mahendran15understanding}. These show that HOG, SIFT, and early layers of CNNs are largely invertible. This apparent inconsistency may be resolved by noting that~\cite{tatu11exploring,szegedy13intriguing,nguyen15deep} require the injection of adversarial noise which is very unlikely to occur in natural images. It is not unlikely that enforcing representation to be sufficiently regular would avoid the issue.


The work by~\cite{nguyen15deep} proposes a second method to generate confounders. In this case, they use genetic programming to create, using a sequence of editing operations, an image that is classified as any desired class by the CNN, while not looking like an instance of any class. The CNN does not have a background class that could be used to reject such images; nonetheless the result is remarkable.



\section{A method for finding the pre-images of a representation}\label{s:method}

This section introduces our method to find pre-images of an image representation. This method will then be applied to the inversion, activation maximization, and caricaturization problems. These are formulated as regularized energy minimization problems where the goal is to find a natural-looking image whose representation has a desired property~\cite{williams86inverting}.
Formally, given a representation function $\Phi : \real^{H\times W \times D} \rightarrow \real^d$ and a reference code $\Phi_0 \in \real^d$, we seek the image\footnote{In the following, the image $\bx$ is assumed to have null mean, as required by most CNN implementations.} $\bx\in\real^{H \times W \times D}$ that minimizes the objective function:
\begin{equation}\label{e:objective}
 \bx^* = \operatornamewithlimits{argmin}_{\bx\in\real^{H \times W \times D}} \mathcal{R}_\alpha(\bx) + \mathcal{R}_{TV^\beta}(\bx) + C \ell(\Phi(\bx),\Phi_0)
\end{equation}
The loss $\ell$ compares the image representation $\Phi(\bx)$ with the target value $\Phi_0$, the two regularizer terms $ \mathcal{R}_\alpha + \mathcal{R}_{TV^\beta} : \real^{H \times W \times D} \rightarrow \real_+$  capture a \emph{natural image prior}, and the constant $C$ trades off loss and regularizers. 

The meaning of minimizing the objective function \eqref{e:objective} depends on the choice of the loss and of the regularizer terms, as discussed below. While these terms contain several parameters, they are designed such that, in practice, all the parameters except $C$ can be fixed for all visualization and representation types.

\subsection{Loss functions}\label{s:loss}


Choosing different loss functions $\ell$ in \autoref{e:objective-jitter} results in different visualizations. In \emph{inversion}, $\ell$ is set to the Euclidean distance:
\begin{equation}\label{e:objective2}
 \ell(\Phi(\bx),\Phi_0) = \frac{\| \Phi(\bx) - \Phi_0 \|^2}{\|\Phi_0\|^2},
\end{equation}
where $\Phi_0 = \Phi(\bx_0)$ is the representation of a target image. Minimizing \eqref{e:objective} results in an image $\bx^*$ that ``resembles'' $\bx_0$ from the viewpoint of the representation.

Sometimes it is interesting to restrict the reconstruction to a subset of the representation components. This is done by introducing a binary \emph{mask} $M$ of the same dimension as $\Phi_0$ and by modifying \autoref{e:objective2} as follows:
\begin{equation}\label{e:objective2p}
 \ell(\Phi(\bx),\Phi_0;M) = \frac{\| (\Phi(\bx) - \Phi_0) \odot M \|^2}{\|\Phi_0 \odot M \|^2},
\end{equation}

In \emph{activation maximization} and \emph{caricaturization}, $\Phi_0 \in \real^d_+$ is treated instead as a weight vector selecting which representation components should be maximally activated. This is obtained by considering the inner product:
\begin{equation}\label{e:objective3}
 \ell(\Phi(\bx),\Phi_0) = - \frac{1}{Z} \langle \Phi(\bx), \Phi_0 \rangle.
\end{equation}
For example, if $\Phi_0= \be_i$ is the indicator vector of the $i$-th component of the representation, minimizing \autoref{e:objective3} maximizes the component $[\Phi(\bx)]_i$. Alternatively, if $\Phi_0$ is set to $\max\{\Phi(\bx_0),0\}$, the minimization of~\autoref{e:objective}  will highlight components that are active in the representation $\Phi(\bx_0)$ of a reference image $\bx_0$, while ignoring the inactive components. 

The choice of the normalization constant $Z$ in activation maximization and caricaturization will be discussed later. Note also that, for the loss \autoref{e:objective3}, there is no need to define a separate mask as this can be pre-multiplied into $\Phi_0$.

\subsection{Regularization}\label{s:regularization}

\begin{figure}
\resizebox{\linewidth}{!}{%
	\includegraphics[width=0.32\linewidth]{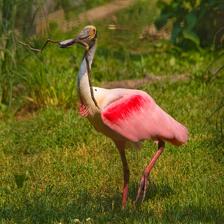}%
	\includegraphics[width=0.32\linewidth]{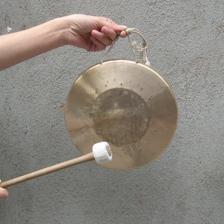}%
	\includegraphics[width=0.32\linewidth]{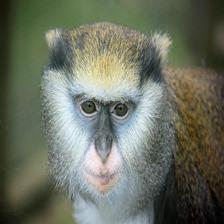}
	}
\resizebox{\linewidth}{!}{%
	\includegraphics[width=0.32\linewidth]{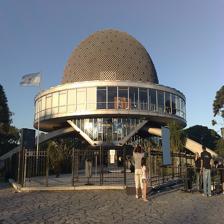}%
	\includegraphics[width=0.32\linewidth]{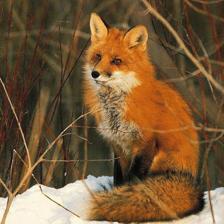}%
	\includegraphics[width=0.32\linewidth]{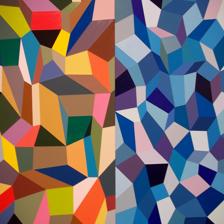}
}
\caption{Input images used in the rest of the paper are shown above. From left to right: Row 1: spoonbill, gong, monkey; Row 2: building, red fox, abstract art.}
	\label{f:allinputimages}
\end{figure}

\begin{figure*}
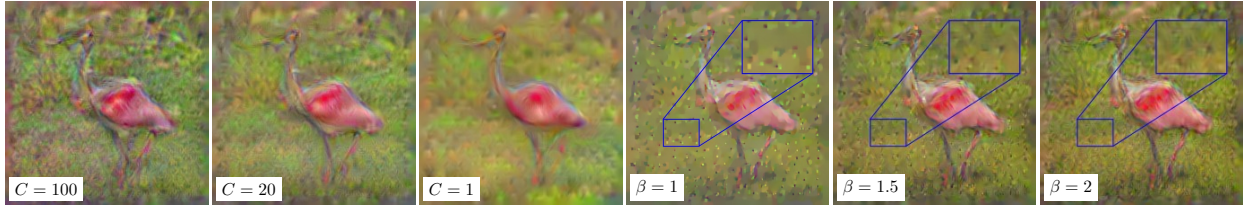

	\resizebox{\linewidth}{!}{
	\centering
	\includelabel{0.25\textwidth}{$C=100$}{{inversion160_imagenet-caffe-alex_ILSVRC2012_val_00000043_layer13-str100}}\ %
	\includelabel{0.25\textwidth}{$C=20$}{{inversion160_imagenet-caffe-alex_ILSVRC2012_val_00000043_layer13-str20}}\ %
	\includelabel{0.25\textwidth}{$C=1$}{{inversion160_imagenet-caffe-alex_ILSVRC2012_val_00000043_layer13-str01}} 
	\includelabel{0.25\textwidth}{$\beta=1$}{{inversion-variants-160_imagenet-caffe-alex_ILSVRC2012_val_00000043_jitter0-beta01zoom}} 
   	\includelabel{0.25\textwidth}{$\beta=1.5$}{{inversion-variants-160_imagenet-caffe-alex_ILSVRC2012_val_00000043_jitter0-beta015zoom}}\ %
	\includelabel{0.25\textwidth}{$\beta=2$}{{inversion-variants-160_imagenet-caffe-alex_ILSVRC2012_val_00000043_jitter0-beta02zoom}}%
	}
	\caption{\emph{Left:} Effect of the data term strength $C$ in inverting a deep representation (the relu3 layer in AlexNet).  Selecting a small value of $C$ results in more regularized reconstructions, which is essential to obtain good results. \emph{Right:} Effect of the TV regularizer $\beta$ exponent; note the spikes for $\beta=1$ (zoomed in the inset). The input image in this case is the ``spoonbill'' image shown in  \autoref{f:allinputimages}.}\label{fig:smoothing}
\end{figure*}
\begin{figure}[t]
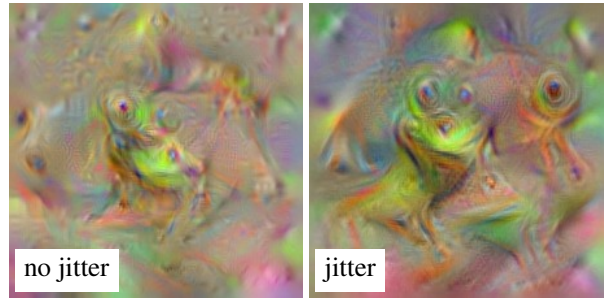

\resizebox{\linewidth}{!}{
\includelabel{0.23\textwidth}{no jitter}{{category-variants-160_imagenet-caffe-alex_treefrog_jitter0-beta02}}\ %
\includelabel{0.23\textwidth}{jitter}{{category-variants-160_imagenet-caffe-alex_treefrog_jitter1-beta02}}
}
\caption{Effect of the jitter regularizer in activation maximization for the ``tree frog'' neuron in the fc8 layer in AlexNet. Jitter helps recover larger and crisper image structures.}\label{fig:jitter}
\end{figure}

Discriminative representations discard a significant amount of low-level image information that is irrelevant to the target task (e.g. image classification).
As this information is nonetheless useful for visualization, we propose to partially recover it by restricting the inversion to the subset of natural images $\mathcal{X}\subset \real^{H\times W \times D}$. This is motivated by the fact that, since representations are applied to natural images, there is comparatively little interest in understanding their behavior outside of this set. However, modeling the set of natural images is a significant challenge in its own right. As a proxy, we propose to regularize the reconstruction by using simple \emph{image priors} implemented as regularizers in~\autoref{e:objective}. We experiment in particular with three such regularizers, discussed next.

\subsubsection{Bounded range}

The first regularizer encourages the intensity of pixels to stay bounded.  This is important for networks that include normalization layers, as in this case arbitrarily rescaling the image range has no effect on the network output. In activation maximization, it is even more important for networks that do \emph{not} include normalization layers, as in this case increasing the image range increases neural activations by the same amount.

 In~\cite{mahendran15understanding} this regularizer was implemented as a soft constraint using the penalty $\|\bx\|_\alpha^\alpha$ for a large value of the exponent $\alpha$.  Here we modify it in several ways. First, for color images we make the term isotropic in RGB space by considering the norm
\begin{equation}\label{e:ennealpha}
  N_\alpha(\bx) =
  \frac{1}{H W B^\alpha}
  \sum_{v=1}^H\sum_{u=1}^W \left(\sum_{k=1}^D \bx(v,u,k)^2\right)^{\frac{\alpha}{2}}
\end{equation}
 where $v$ indexes the image rows, $u$ the image columns, and $k$ the  color channels. By comparison, the norm used in~\cite{mahendran15understanding} is non-isotropic and might slightly bias the reconstruction of colors. 
 
 The term is normalized by the image area $H W$ and by the scalar $B$. This scalar is set to the typical $L^2$ norm of the pixel RGB vector, such that $N_\alpha(\bx) \approx 1$. 
 
 The soft constraint $N_\alpha(\bx)$ is combined with a hard constraint to limit the pixel intensity to be at most $B_+$:
\begin{equation}\label{e:rbound}
	    R_\alpha(\bx) = 
 \begin{cases}
	N_\alpha(\bx), & \forall v,u : \sqrt{\sum_{k} \bx(v,u,k)^2} \leq B_+\\
	+\infty, & \text{otherwise}.
\end{cases}
\end{equation}
While the hard constraint may seem sufficient, in practice it was observed that without soft constraints, pixels tend to saturate in the reconstructions. 
  
\subsubsection{Bounded variation}

The second regularizer is the \emph{total variation} (TV) $\mathcal{R}_\TV(\bx)$ of the image, encouraging reconstructions to consist of piece-wise constant patches. For a discrete image $\bx$, the TV norm is approximated using finite differences as follows:
\begin{multline*}	
 \mathcal{R}_\TV(\bx)
 =
 \frac{1}{HWV^\beta}
 \sum_{uvk}
 \Big(
 \left(\bx(v,u+1,k) - \bx(v,u,k)\right)^2 \\
 +
 \left(\bx(v+1,u,k) - \bx(v,u,k))\right)^2
 \Big)^\frac{\beta}{2}
\end{multline*}
where $\beta=1$. Here the constant $V$ in the normalization coefficient is the typical value of the norm of the gradient in the image.

The standard TV regularizer, obtained for $\beta=1$,  was observed to introduce unwanted ``spikes'' in the reconstruction, as  illustrated in \autoref{fig:smoothing} (right) when inverting a layer of a CNN.
This is a known problem in TV-based image interpolation (see \eg Figure~3 in~\cite{chen2014bi}). The ``spikes'' occur at the locations of the samples because: (1) the TV norm along any path between two samples depends only on the overall amount of intensity change (not on the sharpness of the changes) and (2) integrated on the 2D image, it is optimal to concentrate sharp changes around a boundary with a small perimeter.
Hyper-Laplacian priors with $\beta < 1$ are often used as a better match of the gradient statistics of natural images~\cite{krishnan09fast}, but they only exacerbate this issue.
Instead, we trade off the sharpness of the image with the removal of such artifacts by choosing $\beta > 1$ which, by penalizing large gradients, distributes changes across regions rather than concentrating them at a point or a curve.
We refer to this as the $\TV$ regularizer.
As seen in \autoref{fig:smoothing} (right), the spikes are removed for $\beta = 1.5, 2$ but the image is blurrier than for $\beta = 1$. At the same time, \autoref{fig:smoothing} (left) illustrates the importance of using the $\TV$ regularizer in obtaining clean reconstructions.
 
\subsubsection{Jitter}\label{s:jitter}

The last regularizer, which is inspired by~\cite{mordvintsev15inceptionism:}, has an implicit form and consists of randomly shifting the input image before feeding it to the representation. Namely, we consider the optimization problem 
\begin{equation}\label{e:objective-jitter}
\begin{split}
\bx^* = \operatornamewithlimits{argmin}_{\bx\in\real^{H \times W \times D}} & \mathcal{R}_\alpha(\bx) + \mathcal{R}_{TV^\beta}(\bx) \\
& + C E_\tau[ \ell(\Phi(\operatorname{jitter}(\bx;\tau)),\Phi_0)]
\end{split}
\end{equation}
where $E[\cdot]$ denotes expectation and $\tau=(\tau_1,\tau_2)$ is a discrete random variable uniformly distributed in the set $\{0,\dots,T-1\}^2$, expressing a random horizontal and vertical translation of at most $T-1$ pixels. The $\operatorname{jitter}(\cdot)$ operator translates and crops $\bx$ as follows:
\[
 [\operatorname{jitter}(\bx;\tau)](v,u) = \bx(v+\tau_2,u+\tau_1)
\]
where $1 \leq v \leq H - T + 1$ and $1 \leq u \leq W - T + 1$. The expectation over $\tau$ is not computed explicitly; instead each iteration of SGD samples a new value of $\tau$. Jittering counterbalances the very significant downsampling performed by the earlier layers of deep CNNs, interpolating between pixels in back-propagation. This generally results in crisper pre-images, particularly in the activation maximization problem (\autoref{fig:jitter}). 

\subsubsection{Texture and style regularizers}\label{s:texture}

For completeness, we note that \autoref{e:objective} can also be used to implement the texture synthesis and style transfer visualizations of ~\cite{gatys15texture,gatys15aneural}. One way to do so is to incorporate their texture/style term as an additional regularizer of the form
\begin{equation} \label{e:texture-synthesis-loss}
\mathcal{R}_\mathrm{tex}(\bx)
= 
\sum_{l=1}^L 
w_l
||\psi \circ \Phi_l(\bx) - \psi \circ \Phi_l(\bx_\mathrm{tex}))||_{\mathrm{fro}}^2
\end{equation}
where $\bx_\mathrm{tex}$ is a reference image defining a texture or ``visual style'', $\Psi_l,l=1,\dots,L$ are increasingly deep layers in a CNN, $w_l \geq 0$ weights, and $\psi$ is the \emph{cross-channel correlation} operator
\[
[\psi \circ \Phi_l(\bx)]_{cc'} =
\sum_{uv}
[\Phi_l(\bx)]_{uvc} [\Phi_l(\bx)]_{uvc'}
\]
where $[\Phi_l(\bx)]_{uvc}$ denotes the $c$-th feature channel activation at location $(u,v)$.

The term $\mathcal{R}_\mathrm{tex}(\bx)$ can be used as an objective function in its own right, yielding texture generation, or as a regularizer in the inversion problem, yielding style transfer.

\subsection{Balancing the loss and the regularizers} \label{s:balancing}

One difficulty in implementing a successful image reconstruction algorithm using the formulation of \autoref{e:objective} is to correctly balance the different terms. The loss functions and regularizers are designed in such a manner that, for reasonable reconstruction $\bx$, they have comparable values (around unity). This normalization, though simple, makes a very significant difference. Without it we need to carefully tune parameters across different representation types. Unless otherwise noted, in the experiments we use the values, $C=1$, $\alpha=6$,  $\beta = 2$, $B = 80$, $B_+ = 2B$, and $V = B/6.5$.

\subsection{Optimization}\label{s:optimization}

\begin{algorithm}[t]
\caption{Stochastic gradient descent for pre-image}\label{a:opt}
\begin{algorithmic}[1]
\Require Given the objective function $E(\cdot)$ and the learning rate $\eta_0$
\State $G_0 \leftarrow 0$, $\mu_0 \leftarrow 0$
\State Initialize $\bx_1$ to random noise
\For{$t = 1$ to $T$}
\State $g_t \leftarrow \nabla E(\bx_t)$ (using backprop)
\State $G_{t} \leftarrow \rho G_{t-1} + g_t^2$ (component-wise)
\State $\displaystyle \eta_t \leftarrow \frac{1}{\frac{1}{\eta_0} + \sqrt{G_t}}$ (component-wise)
\State $\mu_{t} \leftarrow \rho \mu_{t-1} - \eta_t g_t$
\State $\bx_{t+1} \leftarrow \Pi_{B_+}\left( \bx_t + \mu_{t} \right)$
\EndFor
\end{algorithmic}
\label{algo:main}
\end{algorithm}

Finding a minimizer of the objective~\eqref{e:objective} may seem difficult as most representations $\Phi$ are strongly non-linear; in particular, deep representations are a composition of several non-linear layers.
Nevertheless, simple gradient descent (GD) procedures have been shown to be very effective in \emph{learning} such models from data, which is arguably an even harder task.
In practice, a variant of GD was found to  result in good reconstructions. 

\paragraph{Algorithm.} The algorithm, whose pseudocode is given in Algorithm~\ref{a:opt}, is a variant of AdaGrad~\cite{duchi11adaptive}. Like in AdaGrad, our algorithm automatically adapts the learning rate of individual components of the vector $\bx_t$ by scaling it by the inverse of the accumulated squared gradient $G_t$. Similarly to AdaDelta~\cite{zeiler12adadelta:}, however, it accumulates gradients only in a short temporal window, using the momentum coefficient $\rho=0.9$. The gradient, scaled by the adaptive learning rate $\eta_t g_t$, is accumulated into a momentum vector $\mu_t$ with the same factor $\rho$. The momentum is then summed to the current reconstruction $\bx_t$ and the result is projected back onto the feasible region $[-B_+, B_+]$. 

Recently, Gatys~\etal~\cite{gatys15aneural,gatys15texture} have used the L-BFGS-B algorithm~\cite{zhu1997algorithm} to optimize their texture/style loss \eqref{e:texture-synthesis-loss}. We found that L-BFGS-B is indeed better than (S)GD for their problem of texture generation, probably due to the particular nature of the term \eqref{e:texture-synthesis-loss}. However, preliminary experiments using L-BFGS-B for inversion did not show a significant benefit, so for simplicity we consider (S)GD-based algorithms in this paper. 

The only parameters of Algorithm~\ref{a:opt} are the initial learning rate $\eta_0$ and the number of iterations $T$. These are discussed next.

\paragraph{Learning rate $\eta_0$.} This parameter can be heuristically set as follows. At the first iteration $G_{0} \approx 0$ and $\eta_1 = 1/(1/\eta_0 + \sqrt{G_0}) \approx \eta_0$; the learning rate $\eta_1$ is approximately equal to the \emph{initial learning rate} $\eta_0$. The value of the step size $\bar \eta_1$ that would minimize the term $R_\alpha(\bx)$ in a single iteration (ignoring momentum) is obtained by solving the equation
$
  0 \approx \bx_2 = \bx_1 - \bar \eta_1 \nabla R_\alpha(\bx_1).
$
Assuming that all pixels in $\bx_1$ have intensity equal to the parameter $B$ introduced above, one then obtains the condition
$ 0 = B - \bar \eta_1 \alpha/B$, so that $\bar \eta_1 = B^2/\alpha$. The initial learning rate $\eta_0$ is set to a hundredth of this value:
$
  \eta_0 = 0.01 \, \bar \eta_1 =  0.01\,B^2/\alpha.
$

\paragraph{Number of iterations $T$.} Algorithm~\ref{a:opt} is run for $T=300$ iterations. When jittering is used as a regularizer, we found it beneficial to eventually disable it and run the algorithm for a further 50 iterations, after reducing the learning rate tenfold. This fine tuning does not change the results qualitatively, but for inversion it slightly improves the reconstruction error; thus it is not applied in caricaturization and activation maximization.
%


The cost of running Algorithm~\ref{a:opt} is dominated by the cost of computing the derivative of the representation function, usually by back-propagation in a deep neural network. By comparison, the cost of computing the derivative of the regularizers and the cost of the gradient update are negligible. This also means that the algorithm runs faster for shallower representations and slower for deeper ones; on a CPU, it may in practice take only a few seconds to visualize shallow layers in a deep network and a few minutes for deep ones. GPUs can accelerate the algorithm by an order of magnitude or more. Another simple speedup is to stop the algorithm earlier; here using 300-350 iterations is a conservative choice that works for all representation types and visualizations we tested.

\section{Representations}\label{s:representations}

%
%

\begin{table}
\renewcommand{\arraystretch}{1.1}
\newcolumntype{Y}{>{\centering\arraybackslash}X}
\newcolumntype{Z}{>{\raggedright\arraybackslash}X}
\tabcolsep=0.05cm
\begin{tabularx}{0.48\textwidth}{|ZYY|ZYY|ZYY|}
	\hline
	\multicolumn{3}{|c}{AlexNet} & \multicolumn{3}{|c}{VGG-M} & \multicolumn{3}{|c|}{VGG-VD-16} \\
	\hline
	name & size & stride & name & size & stride & name & size & stride \\
	\hline
	conv1 & 11  & 4  & conv1 & 7   & 2  & conv1\_1 & 3   & 1  \\
	relu1 & 11  & 4  & relu1 & 7   & 2  & relu1\_1 & 3   & 1  \\
	&     &    &       &     &    & conv1\_2 & 5   & 1  \\
	&     &    &       &     &    & relu1\_2 & 5   & 1  \\
	norm1 & 11  & 4  & norm1 & 7   & 2  &          &     &    \\
	pool1 & 19  & 8  & pool1 & 11  & 4  & pool1    & 6   & 2  \\
	\hline
	conv2 & 51  & 8  & conv2 & 27  & 8  & conv2\_1 & 10  & 2  \\
	relu2 & 51  & 8  & relu2 & 27  & 8  & relu2\_1 & 10  & 2  \\
	&     &    &       &     &    & conv2\_2 & 14  & 2  \\
	&     &    &       &     &    & relu2\_2 & 14  & 2  \\
	norm2 & 51  & 8  & norm2 & 27  & 8  &          &     &    \\
	pool2 & 67  & 16 & pool2 & 43  & 16 & pool2    & 16  & 4  \\
	\hline
	conv3 & 99  & 16 & conv3 & 75  & 16 & conv3\_1 & 24  & 4  \\
	relu3 & 99  & 16 & relu3 & 75  & 16 & relu3\_1 & 24  & 4  \\
	&     &    &       &     &    & conv3\_2 & 32  & 4  \\
	&     &    &       &     &    & relu3\_2 & 32  & 4  \\
	&     &    &       &     &    & conv3\_3 & 40  & 4  \\
	&     &    &       &     &    & relu3\_3 & 40  & 4  \\
	&     &    &       &     &    & pool3    & 44  & 8  \\
	\hline	      
	conv4 & 131 & 16 & conv4 & 107 & 16 & conv4\_1 & 60  & 8  \\ 
	relu4 & 131 & 16 & relu4 & 107 & 16 & relu4\_1 & 60  & 8  \\
	&     &    &       &     &    & conv4\_2 & 76  & 8  \\
	&     &    &       &     &    & relu4\_2 & 76  & 8  \\
	&     &    &       &     &    & conv4\_3 & 92  & 8  \\
	&     &    &       &     &    & relu4\_3 & 92  & 8  \\
	&     &    &       &     &    & pool4    & 100 & 16 \\
	\hline
	conv5 & 163 & 16 & conv5 & 139 & 16 & conv5\_1 & 132 & 16 \\
	relu5 & 163 & 16 & relu5 & 139 & 16 & relu5\_1 & 132 & 16 \\ 
	&     &    &       &     &    & conv5\_2 & 164 & 16 \\
	&     &    &       &     &    & relu5\_2 & 164 & 16 \\
	&     &    &       &     &    & conv5\_3 & 196 & 16 \\
	&     &    &       &     &    & relu5\_3 & 196 & 16 \\
	pool5 & 195 & 32 & pool5 & 171 & 32 & pool5    & 212 & 32 \\
	\hline
	fc6   & 355 & 32 & fc6   & 331 & 32 & fc6      & 404 & 32 \\
	relu6 & 355 & 32 & relu6 & 331 & 32 & relu6    & 404 & 32 \\
	fc7   & 355 & 32 & fc7   & 331 & 32 & fc7      & 404 & 32 \\
	relu7 & 355 & 32 & relu7 & 331 & 32 & relu7    & 404 & 32 \\
	fc8   & 355 & 32 & fc8   & 331 & 32 & fc8      & 404 & 32 \\ 
	prob  & 355 & 32 & prob  & 331 & 32 & prob     & 404 & 32 \\
	\hline
\end{tabularx}
	\caption{{\bf CNN architectures.} Structure of the AlexNet, VGG-M and VGG-VD-16 CNNs, including the layer names, the receptive field sizes (size), and the strides (stride) between feature samples, both in pixels. Note that, due to down-sampling and padding, the receptive field size can be larger than the size of the input image.}\label{f:allnetworks}
\end{table}

In this section, the image representations studied in the paper - dense SIFT, HOG, and several reference deep CNNs, are described. It is also shown how DSIFT and HOG can be implemented in a standard CNN framework, which simplifies the computation of their derivatives as required by the algorithm of~\autoref{s:optimization}.
 
\subsection{Classical representations}\label{s:shallow-networks}

The \emph{histograms of oriented gradients} are probably the best known family of ``classical'' computer vision features popularized by Lowe in~\cite{lowe99object} with the SIFT descriptor. Here we consider two densely-sampled versions~\cite{nowak06sampling}, namely DSIFT (Dense SIFT) and HOG~\cite{dalal05histograms}. In the remainder of this section these two representations are reformulated as CNNs. This clarifies the relationship between SIFT, HOG, and CNNs in general and helps implement them in standard CNN toolboxes for experimentation. The DSIFT and HOG implementations in the VLFeat library~\cite{vedaldi07open} are used as numerical references. These are equivalent to Lowe's~\cite{lowe99object} SIFT and the DPM V5~HOG~\cite{lsvm-pami,voc-release5}. 

SIFT and HOG involve: computing and binning image gradients, pooling binned gradients into cell histograms, grouping cells into blocks, and normalizing the blocks.
Let us denote by $\bg$ the image gradient at a given pixel and consider binning this into one of $K$ orientations (where $K=8$ for SIFT and $K=18$ for HOG).
This can be obtained in two steps: directional filtering and non-linear activation.
The $k^{th}$ directional filter is $G_k = u_{1k} G_x + u_{2k } G_y$ where
\[
\bu_k = \begin{bmatrix} \cos \frac{2\pi k}{K} \\ \sin \frac{2\pi k}{K} \end{bmatrix},
\quad
 G_x = \begin{bmatrix} 0 & 0 & 0 \\ -1 & 0 & 1 \\ 0 & 0 & 0 \end{bmatrix},
\quad
 G_y = G_x^\top.
\]
The output of a directional filter is the projection $\langle \bg, \bu_k \rangle$ of the gradient along direction $\bu_k$. This is combined with a non-linear activation function to assign gradients to histogram elements $h_k$. DSIFT uses bilinear orientation assignment, given by
\[
  h_k= \|\bg\| 
  \max\left\{0, 1 - \frac{K}{2\pi} \cos^{-1} \frac{\langle \bg, \bu_k \rangle}{\|\bg\|} \right\},
\]
whereas HOG (in the DPM V5 variant) uses hard assignment $h_k = \|\bg\| \mathbf{1}\left[\langle \bg, \bu_k \rangle > \|\bg\| \cos\pi/K \right]$.
Filtering is a standard CNN operation but these activation functions are not.
While their implementation is simple, an interesting alternative is to approximate bilinear orientation assignment by using the activation function:
\begin{align*}
  h_k 
  &\approx \|\bg\| 
  \max\left\{0, \frac{1}{1-a} \frac{\langle \bg, \bu_k \rangle }{\|\bg\|} - \frac{a}{1-a}\right\}
  \\
  &\propto \max\left\{0, \langle \bg, \bu_k \rangle - a\|\bg\| \right\},
  \quad a = \cos 2\pi/K.
\end{align*}
This activation function is the standard ReLU operator modified to account for the norm-dependent offset $a\|\bg\|$. While the latter term is still non-standard, this indicates that a close approximation of binning can be achieved in standard CNN architectures.

The next step is to pool the binned gradients into cell histograms using bilinear spatial pooling, followed by extracting blocks of $2\times 2$ (HOG) or $4 \times 4$ (SIFT) cells.
Both operations can be implemented by banks of linear filters.
Cell blocks are then $l^2$ normalized, which is a special case of the standard local response normalization layer.
For HOG, blocks are further decomposed back into cells, which requires another filter bank.
Finally, the descriptor values are clamped from above by applying $y = \min\{x,0.2\}$ to each component, which can be reduced to a combination of linear and ReLU layers.

The conclusion is that approximations to DSIFT and HOG can be implemented with conventional CNN components plus the non-conventional gradient norm offset.
However, all the filters involved are much sparser and simpler than the generic 3D filters in learned CNNs. 
Nonetheless, in the rest of the paper we will use exact CNN equivalents of DSIFT and HOG, using modified or additional CNN components as needed.%
\footnote{This requires addressing a few more subtleties. Please see files \texttt{dsift\_net.m} and \texttt{hog\_net.m} for details.}
These CNNs are numerically indistinguishable from the VLFeat reference implementations, but, true to their CNN nature, allow computing the feature derivatives as required by the algorithm of \autoref{s:optimization}.

\subsection{Deep convolutional neural networks}\label{s:deep-networks}

The first CNN model considered in this paper is {\bf AlexNet}. Due to its popularity, we use the implementation that ships with the Caffe framework~\cite{jia13caffe}, which closely reproduces the original network by Krizhevsky \etal~\cite{krizhevsky12imagenet}.  Occasionally, we also consider the {\bf CaffeNet}, a network similar to AlexNet that also comes with Caffe.
This and many other similar networks alternate the following computational building blocks: linear convolution, ReLU, spatial max-pooling, and local response normalization.
Each such block takes as input a $d$-dimensional image and produces as output a $k$-dimensional one.
Blocks can additionally pad the image (with zeros for the convolutional blocks and with $-\infty$ for max pooling) or subsample the data.
The last several layers are deemed ``fully connected'' as the support of the linear filters coincides with the size of the image; however, they are equivalent to filtering layers in all other respects. \autoref{f:allnetworks} (left) details the structure of AlexNet.

The second network is the {\bf VGG-M} model from~\cite{chatfield14return}.
The structure of VGG-M (\autoref{f:allnetworks} -- middle) is very similar to AlexNet, with the following differences: it includes a significantly larger number of filters in the different layers, filters at the beginning of the network are smaller, and filter strides (subsampling) is reduced. While the network is slower than AlexNet, it also performs much better on the ImageNet ILSVRC 2012 data.

The last network is the {\bf VGG-VD-16} model from~\cite{simonyan14deep}. 
VGG-VD-16 is also similar to AlexNet, but with more substantial changes compared to VGG-M  (\autoref{f:allnetworks} -- right). Filters are very narrow ($3\times 3$) and very densely sampled. There are no normalization layers. Most importantly, the network contains many more intermediate convolutional layers. The resulting model is very slow, but very powerful.

All pre-trained models are implemented in the MatConvNet framework and are publicly available at~\url{http://www.vlfeat.org/matconvnet/pretrained}.

\begin{figure*}[ht!]\centering
\newcommand{\tri}{trim={0 0 {.5\width} {.5\height}},clip}
\puti{(a) Orig.}{\includegraphics[width=0.15\textwidth]{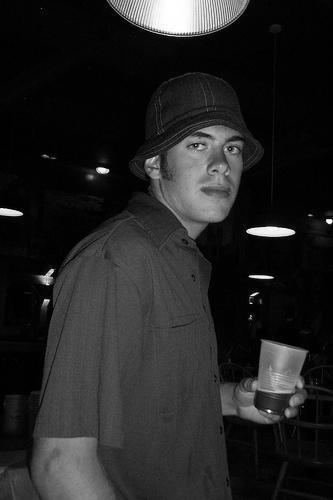}}%
\puti{(b) HOG}{\includegraphics[width=0.15\textwidth]{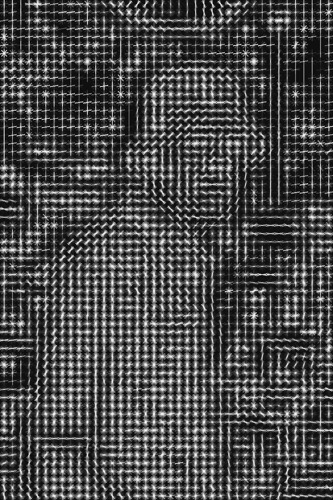}}%
\puti{(c) HOGgle~\cite{vondrick13hoggles:}}{\includegraphics[width=0.15\textwidth]{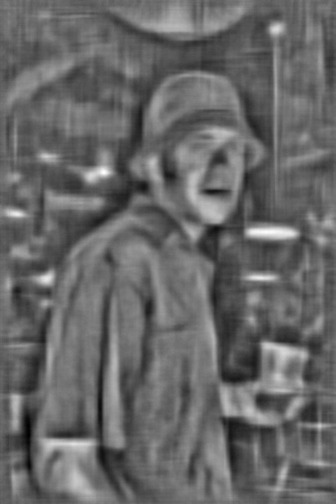}}%
\puti{(d) $\text{HOG}^{-1}$}{\includegraphics[width=0.15\textwidth]{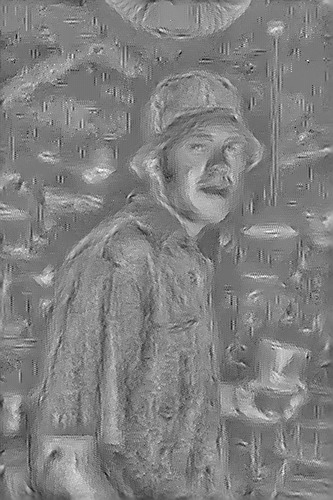}}%
\puti{(e) $\text{HOGb}^{-1}$}{\includegraphics[width=0.15\textwidth]{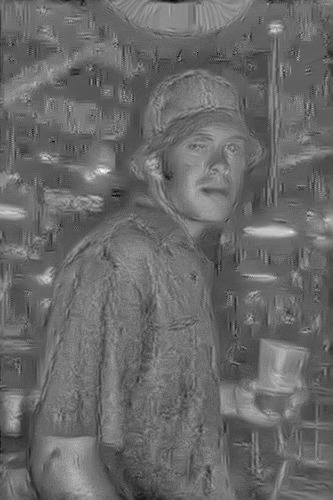}}%
\puti{(f) $\text{DSIFT}^{-1}$}{\includegraphics[width=0.15\textwidth]{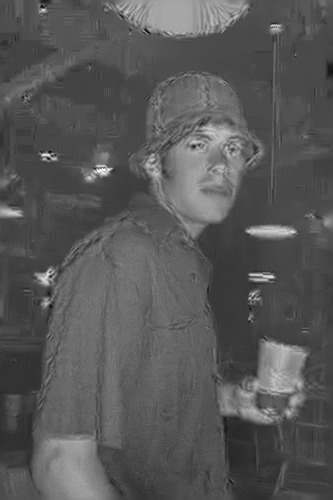}}
{\adjincludegraphics[width=0.15\textwidth,trim={150pt 300pt 80pt 100pt},clip]{pics_hoggle_hoggle-orig-1-bw}}%
{\adjincludegraphics[width=0.15\textwidth,trim={150pt 300pt 80pt 100pt},clip]{pics_hoggle_hoggle-orig-1-hog}}%
{\adjincludegraphics[width=0.15\textwidth,trim={150pt 300pt 80pt 100pt},clip]{inversion160_ihog_hoggle-orig-1_res}}%
{\adjincludegraphics[width=0.15\textwidth,trim={150pt 300pt 80pt 100pt},clip]{inversion160_hog_hoggle-orig-1_res}}%
{\adjincludegraphics[width=0.15\textwidth,trim={150pt 300pt 80pt 100pt},clip]{inversion160_hogb_hoggle-orig-1_res}}%
{\adjincludegraphics[width=0.15\textwidth,trim={150pt 300pt 80pt 100pt},clip]{inversion160_dsift_hoggle-orig-1_res}}
\caption{Reconstruction quality of different representation inversion methods, applied to HOG and DSIFT. HOGb denotes HOG with bilinear orientation assignments. This image is best viewed on screen.}\label{f:hoggles}
\end{figure*}

\section{Visualization by inversion}\label{s:results-inversion}

The experiments in this section apply the visualization by inversion method to both classical (\autoref{s:results-inversion-classical}) and CNN (\autoref{s:result-inversion-cnn}) representations. As detailed in \autoref{s:loss}, for inversion, the objective function \eqref{e:objective} is set up to minimize the $L^2$ distance \eqref{e:objective2} between the representation $\Phi(\bx)$ of the reconstructed image $\bx$ and the representation $\Phi_0=\Phi(\bx_0)$ of a reference image $\bx_0$. 

Importantly, the optimization starts by initializing the reconstructed image to random i.i.d. noise such that \emph{the only information available to the algorithm is the code $\Phi_0$}. When started from different random initializations, the algorithm is expected to produce different reconstructions. This is partly due to the local nature of the optimization, but more fundamentally to the fact that representations are designed to be invariant to nuisance factors. Hence, images with irrelevant differences should have the same representation and should be considered equally good reconstruction targets. In fact, it is by observing the differences between such reconstructions that we can obtain insights into the nature of the representation invariances.

Due to their intuitive nature, it is not immediately obvious how visualizations should be assessed quantitatively. Here we do so from multiple angles. The first is to test whether the algorithm successfully attains its goal of reconstructing an image $\bx$ that has the desired representation $\Phi(\bx)=\Phi_0$. In \autoref{s:results-inversion-classical} and \autoref{s:result-inversion-cnn} this is tested in terms of the relative reconstruction error of~\autoref{e:objective2}. Furthermore, in \autoref{s:result-inversion-cnn} it is also verified for CNNs whether the reconstructed and original representations have the same ``meaning'', in the sense that they are mapped to the same class label. Note that such tests assess the reconstruction quality in \emph{feature space} rather than in \emph{image space}. This is an important point: as noted above, we are not interested in recovering an image $\bx$ which is perceptually similar to the reference image $\bx_0$; rather, in order to study the invariances of the representation $\Phi$, we would like to recover an image $\bx$ that differs from $\bx_0$ but has the same representation. Measuring the difference in feature space is therefore appropriate. 

Finally, the effect of regularization is assessed empirically, via human assessment, to check whether the proposed notion of naturalness does in fact improve the interpretability of the visualizations.

\subsection{Inverting classical representations: SIFT and HOG}\label{s:results-inversion-classical}

\begin{table}
\centering
\begin{tabular}{|c|cc|c|c|}
\hline
descriptors & HOG & HOG & HOGb & DSIFT \\
method & HOGgle & our & our & our \\
\hline
error (\%) & ${60.1}$ & ${36.6}$ & ${11.6}$ & ${9.4}$\\[-0.5em]
& \tiny $\pm {2.3}$ & \tiny $\pm {3.4}$ & \tiny $\pm {0.9}$ & \tiny $\pm {1.7}$\\
\hline
\end{tabular}
\vspace{0.5em}
\caption{Average reconstruction error of different representation inversion methods, applied to HOG and DSIFT. HOGb denotes HOG with bilinear orientation assignments. The error bars show the 95\% confidence interval for the mean.}\label{t:hog-errors}
\end{table}

In this section the visualization by inversion method is applied to the HOG and DSIFT representations.

\subsubsection{Implementation details}

The parameter $C$ in~\autoref{e:objective}, trading off regularization and feature reconstruction fidelity, is set to 100 unless noted otherwise. Jitter is not used and the other parameters are set as stated in~\autoref{s:balancing}. HOG and DSIFT cell sizes are set to 8 pixels.


\subsubsection{Reconstruction quality}

Based on the discussion above, the reconstruction quality is assessed by reporting the normalized reconstruction error \eqref{e:objective2p}, averaged over the first 100 images in the ILSVRC 2012 challenge validation images~\cite{ILSVRC15}. The closest alternative to our inversion method is HOGgle, a technique introduced by Vondrick~\etal~\cite{vondrick13hoggles:} for visualizing HOG features. The HOGgle code is publicly available from the authors' website and is used throughout these experiments. HOGgle is pre-trained to invert the UoCTTI variant of HOG, which is numerically equivalent to the CNN-HOG network of \autoref{s:representations}, which allows us to compare algorithms directly.


Compared to our method, HOGgle is faster (2-3s vs. 60s on the same CPU) but not as accurate, as is apparent both qualitatively (\autoref{f:hoggles}.c vs. d) and quantitatively (60.1\% vs. 36.6\% reconstruction error, see \autoref{t:hog-errors}). Notably, Vondrick~\etal did propose a direct optimization method similar to~\eqref{e:objective}, but found that it did not perform better than HOGgle. This demonstrates the importance of the choice of regularizer and of the ability to compute the derivative of the representation analytically in order to implement  optimization effectively.
%
In terms of speed, an advantage of optimizing~\eqref{e:objective} is that it can be switched to use GPU code immediately given the underlying CNN framework; doing so results in a ten-fold speed-up. 


\subsubsection{Representation comparison}

Different representations are easier or harder to invert.
For example, modifying HOG to use bilinear gradient orientation assignments as in SIFT \mbox{(\autoref{s:representations})} significantly reduces the reconstruction error (from 36.6\% down to 11.5\%) and improves the reconstruction quality (\autoref{f:hoggles}.e).
More remarkable are the reconstructions obtained by inverting DSIFT: they are quantitatively similar to HOG with bilinear orientation assignment, but produce significantly more detailed images (\autoref{f:hoggles}.f).
Since HOG uses a finer quantization of the gradient compared to SIFT but otherwise the same cell size and sampling, this result can be imputed to the stronger normalization in HOG that evidently discards more visual information than in SIFT.

\subsection{Inverting CNNs}\label{s:result-inversion-cnn}

In this section the visualization by inversion method is applied to representative CNNs: AlexNet, VGG-M, and VGG-VD-16.

\subsubsection{Implementation details}

The jitter amount $T$ is set to the integer closest to a quarter of the \emph{stride} of the representation; the stride is the step in the receptive field of representation components when stepping through spatial locations. Its value is given in \autoref{f:allnetworks}. The other parameters are set as stated in~\autoref{s:balancing}.

The parameter $C$ in~\autoref{e:objective} is set to one of 1, 20, 100 or 300. Based on the analysis below, unless otherwise specified, visualizations use the following values: For AlexNet and VGG-M, we choose $C=300$ up to relu3, $C=100$ up to relu4, $C=20$ up to relu5, and $C=1$ for the remaining layers. For VGG-VD we use $C=300$ up to conv4\_3, $C=100$ for conv5\_1, $C=20$ up to conv5\_3, and $C=1$ onwards.

\begin{figure*}[t]
\centering
\resizebox{\textwidth}{!}{%
\includegraphics[width=0.3\linewidth]{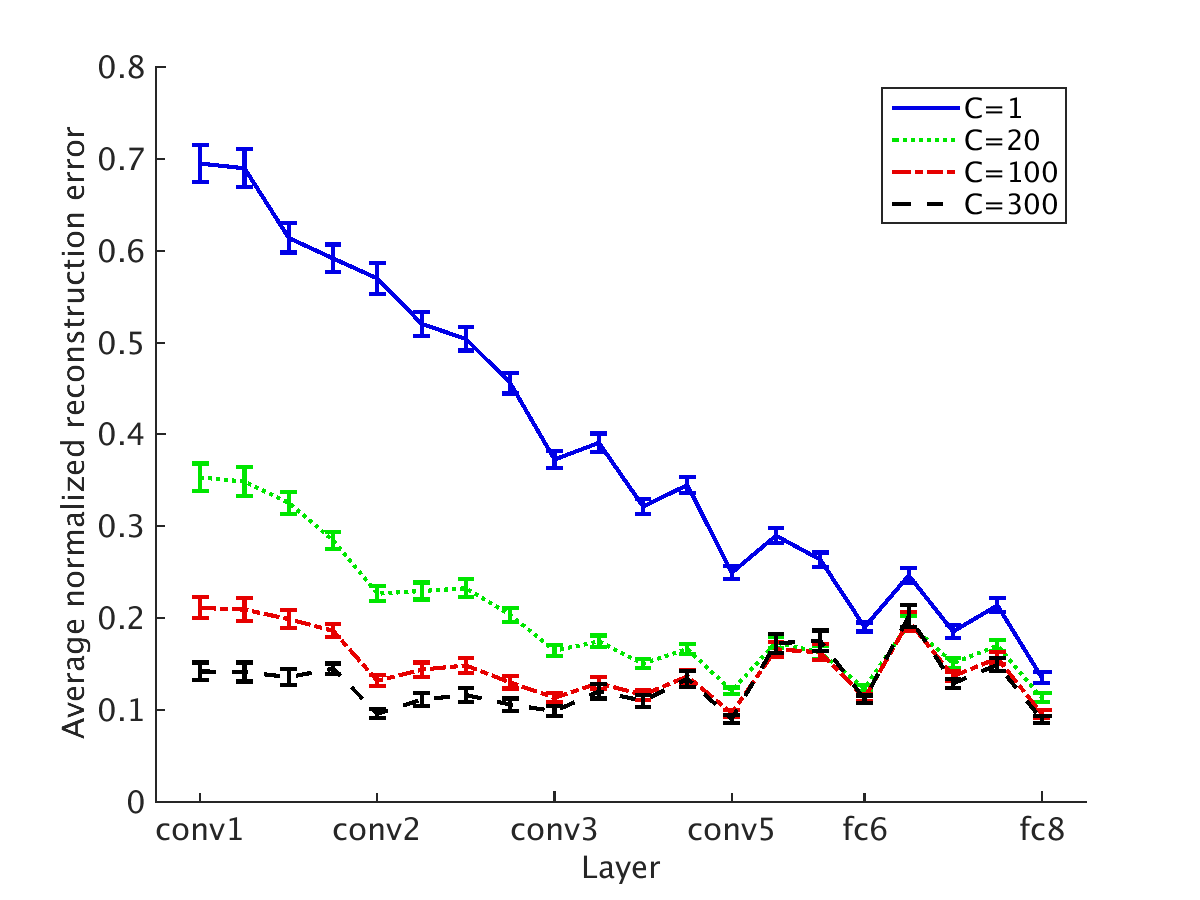}%
\includegraphics[width=0.3\linewidth]{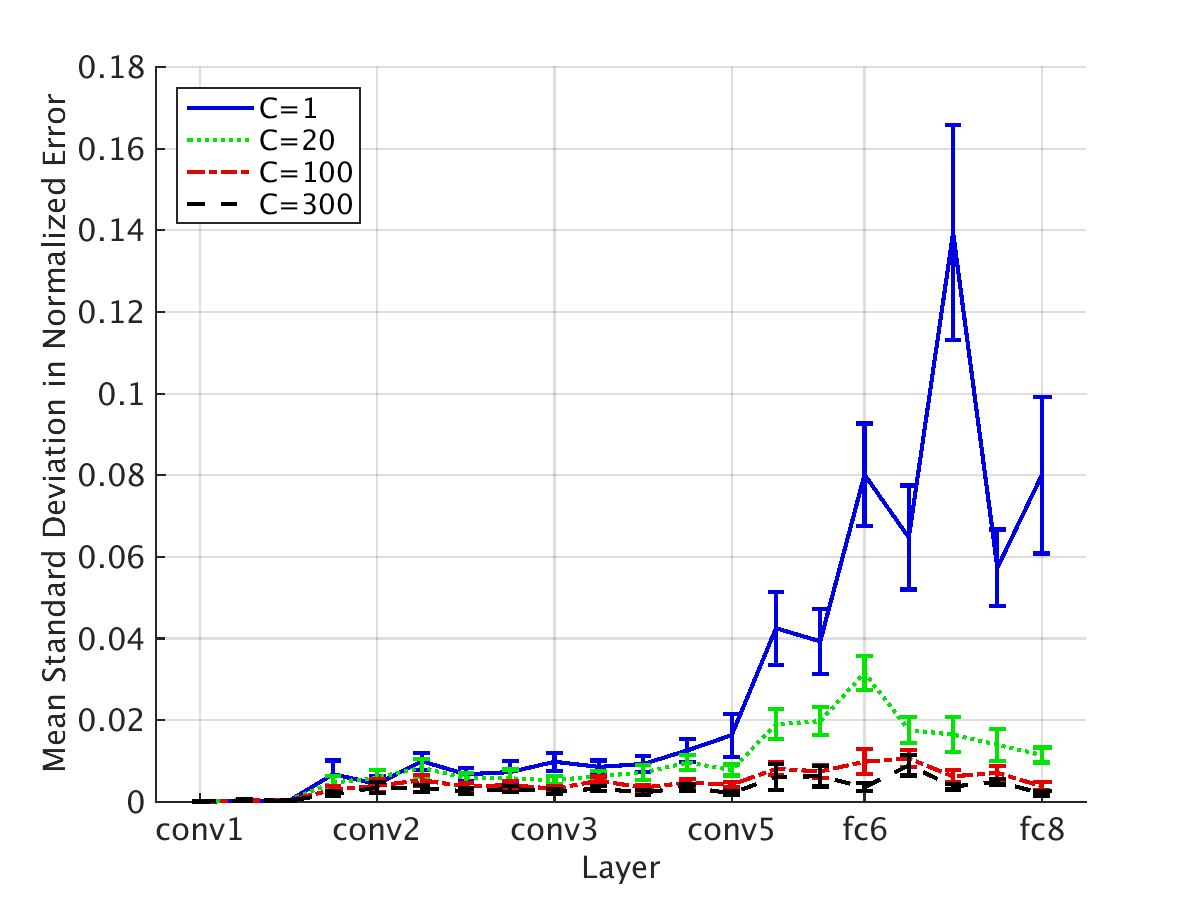}%
\includegraphics[width=0.3\linewidth]{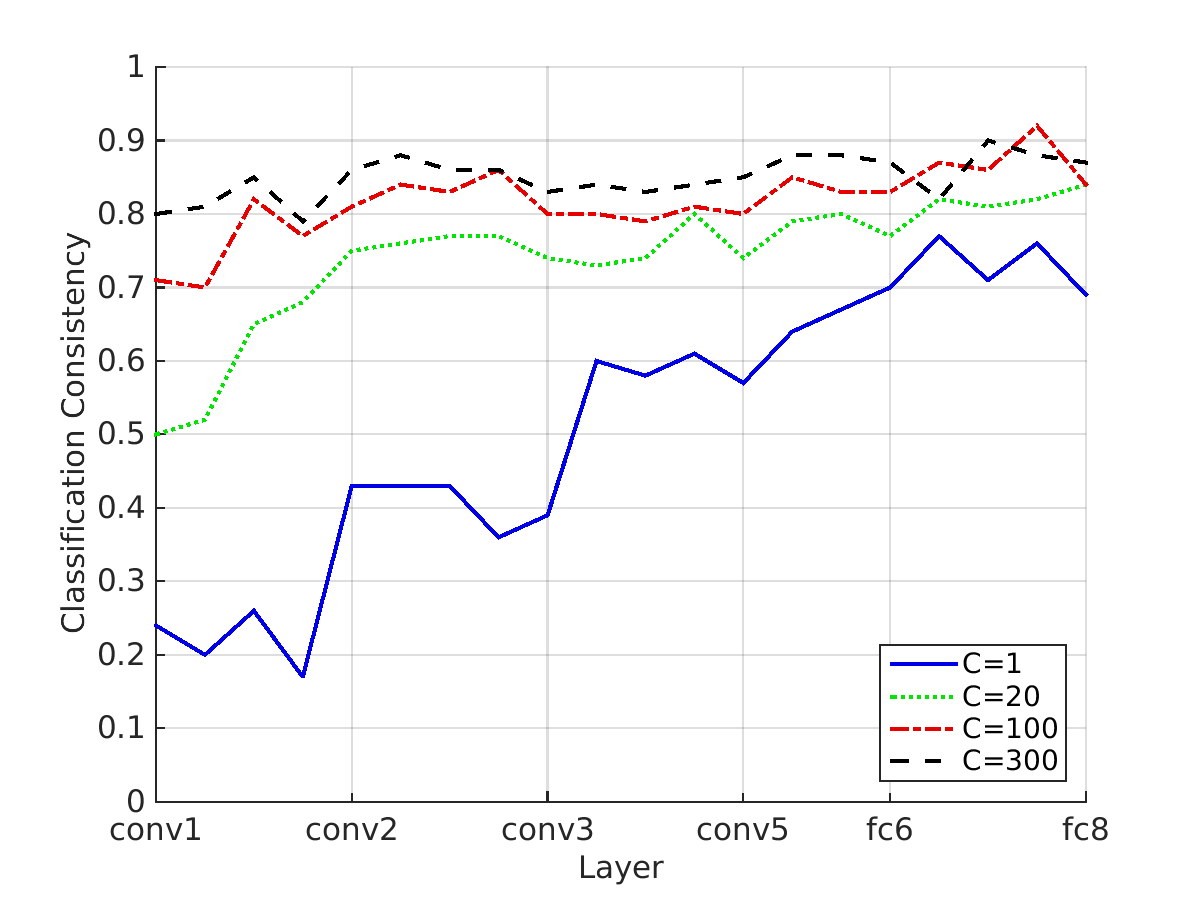}%
}
\caption{\emph{Quality of CNN inversions.} The plots report three reconstruction quality indices for the CNN inversion method applied to different layers of AlexNet and for different reconstruction fidelity strengths $C$. The left plot shows the average reconstruction error averaged using 100 ImageNet ILSVRC validation images as reference images (the error bars show the 95\% confidence interval for the mean). The middle plot reports the standard deviation of the reconstruction error obtained from 24 different random initializations per reference image, and then averaged over 24 such images. The right plot reports the percentage of reconstructions that are associated by the CNN to the same class as their reference image.}
\label{f:cnn-errors}
\end{figure*}

\subsubsection{Reconstruction quality}

The reconstruction accuracy is assessed in three ways: reconstruction error,  consistency with respect to different random initializations, and classification consistency.

\paragraph{Reconstruction error.} Similar to~\autoref{s:results-inversion-classical}, the reconstruction error~\eqref{e:objective2p} is averaged over the first 100 images in the ILSVRC 2012 challenge validation images~\cite{ILSVRC15} (these images were not used to train the CNNs). The experiment is repeated for all the layers of AlexNet and for different values of the parameter $C$ to assess its effect. The resulting average errors are reported in \autoref{f:cnn-errors} (left panel).

CNNs such as AlexNet are significantly larger and deeper than the CNN implementations of HOG and DSIFT. Therefore, it seems that the inversion problem should be considerably harder for them. Instead, comparing the results in \autoref{f:cnn-errors} to the ones in \autoref{t:hog-errors} indicates that CNNs are, in fact, not much more difficult to invert than HOG. In particular, for a sufficiently large value of $C$, the reconstruction error can be maintained in the range $10$--$20\%$, including for the deepest layers. Therefore, the non-linearities in the CNN seem to be rather benign, which could explain why SGD can learn these models successfully. Using a stronger regularization (small $C$) significantly deteriorates the quality of the reconstructions from earlier layers of the network. At the same time, it has little to no effect on the reconstruction quality of deeper layers. Since, as verified below, a strong regularization significantly improves the interpretability of resulting pre-images, it should be used for these layers.

\paragraph{Consistency of multiple pre-images.} As explained earlier, different random initializations are expected to result in different reconstructions. However, this diversity should reflect genuine representation ambiguities and invariances rather than the inability of the local optimization method to escape bad local optima. To verify that this is the case, the standard deviation of the reconstruction error~\eqref{e:objective2p} is computed from 24 different reconstructions obtained from the same reference image and 24 different initializations. The experiment is repeated using as reference the first 24 images in the ILSVRC 2012 validation dataset and the average standard deviation of the reconstruction errors is reported in~\autoref{f:cnn-errors} (middle panel). This figure shows that, for the values of $C$ except $C=1$, all pre-images have very similar reconstruction errors, with standard deviation of around 0.02 or less. Thus in all but the very deep layers all pre-images can be treated as equally good from the viewpoint of reconstruction error. In the next paragraph, we show that even for very deep layers, pre-images are substantially equivalent from the viewpoint of classification consistency.

\paragraph{Classification consistency.} One question that may arise is whether a reconstruction error of $20\%$, or even $10\%$, is sufficiently small to validate the visualizations. To answer this question, \autoref{f:cnn-errors} (right panel) reports the \emph{classification consistency} of the reconstructions. Here ``classification consistency'' is the fraction of reconstructed pre-images $\bx$ that the CNN associates with the same class label as the reference image $\bx_0$. This value, which would be equal to 1 for perfect reconstructions, measures whether imperfections in the reconstruction process are small enough to not affect the ``meaning'' of the image from the viewpoint of the CNN.

As it may be expected, classification consistency results show a trend similar to the reconstruction error, where better reconstructions are obtained for small amounts of regularization for the shallower layers, whereas deep layers can afford much stronger regularization. It is interesting to note that even visually odd inversions of deep layers such as those shown in \autoref{f:splash} or \autoref{f:inversion-alexnet} are classification consistent with their reference image, demonstrating the high degree of invariance in such layers. Finally, we note that, by choosing the correct amount of regularization, the classification consistency of the reconstructions can be kept above 0.8 in most cases, validating the visualizations.

\begin{figure}[t]
\centering
\includegraphics[width=.8\linewidth]{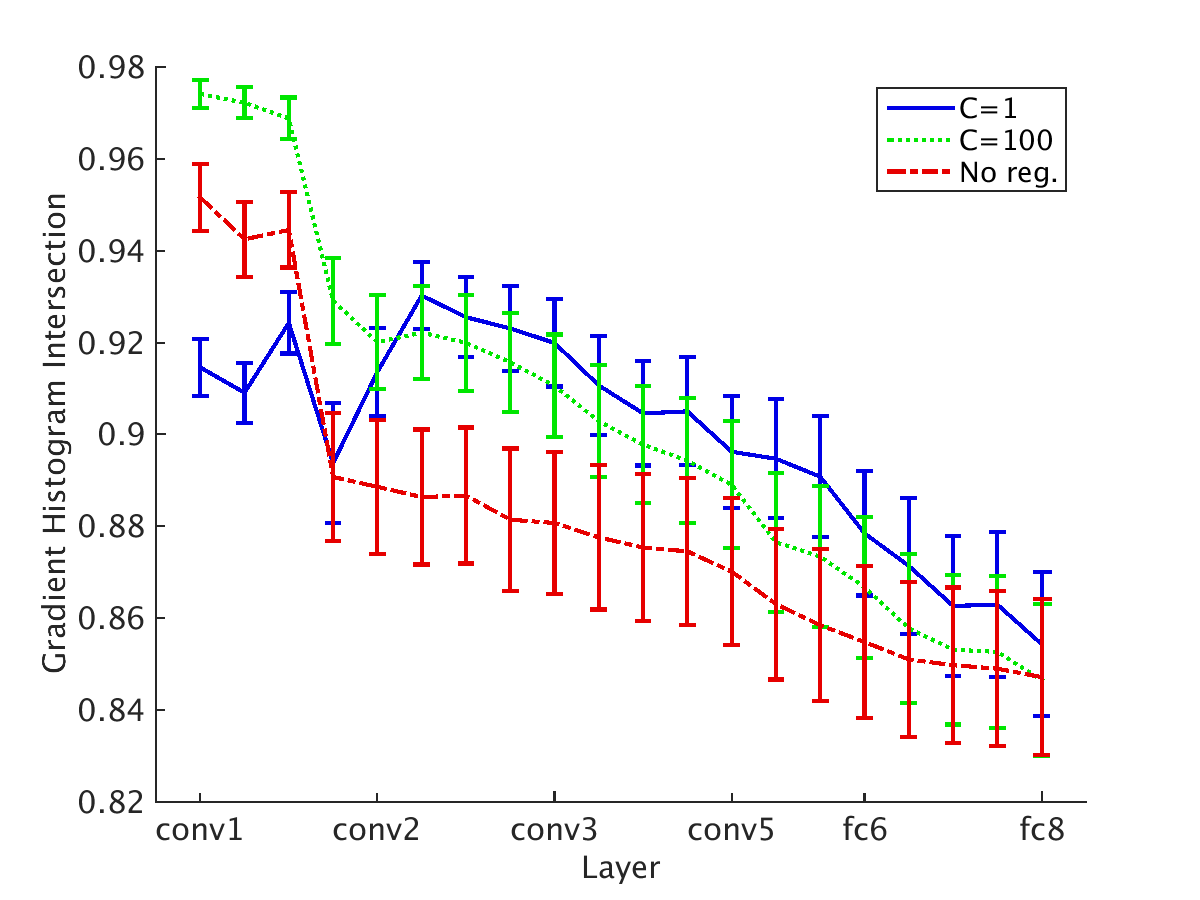}
\caption{Mean histogram intersection similarity for the gradient statistics of the reference image $\bx_0$ and the reconstructed pre-image $\bx$ for different layers of AlexNet and values of the parameter $C$ (only a few such values are reported to reduce clutter). Error bars show the 95\% confidence interval of the mean histogram intersection similarity.}
\label{f:naturalness}
\end{figure}

\begin{figure}[t]
	\centering
	\includegraphics[width=0.8\linewidth]{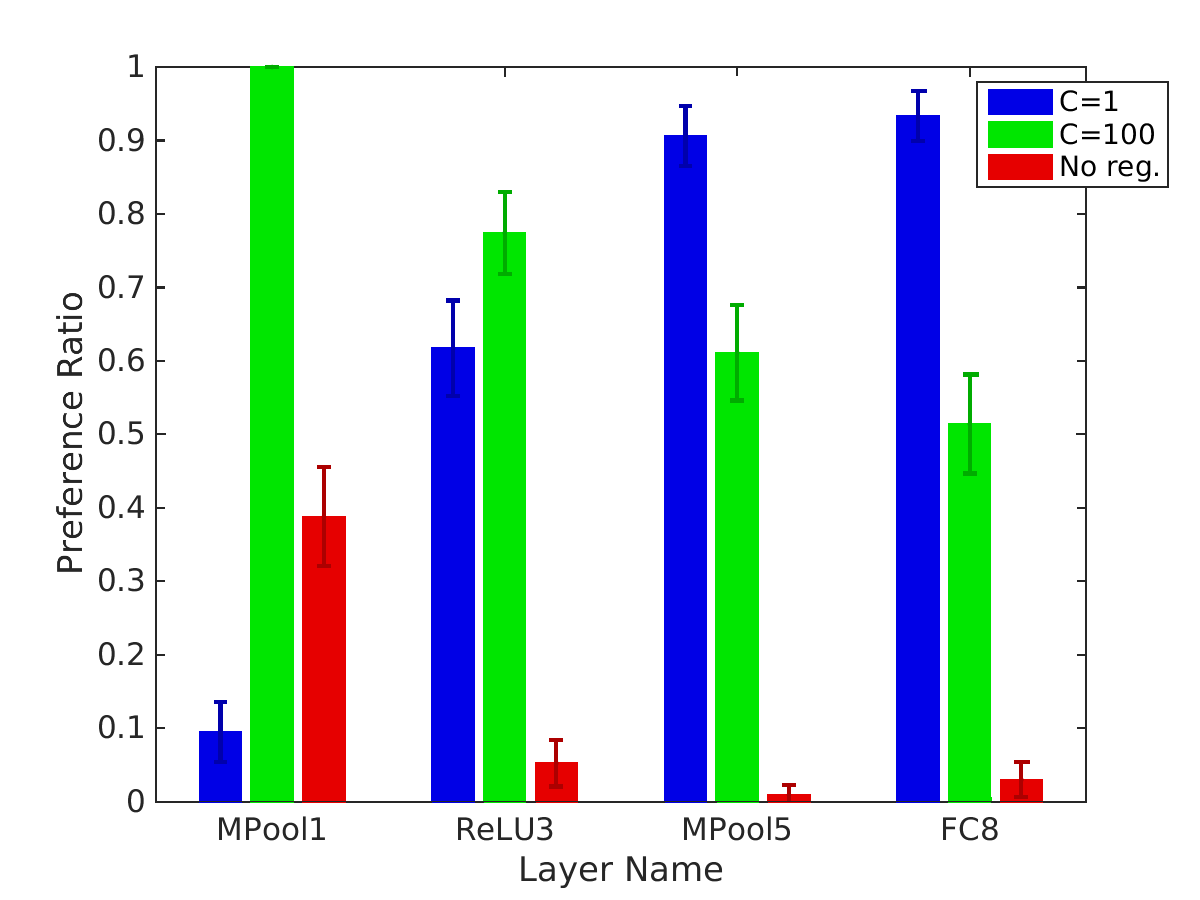}
	\caption{Fraction of times a certain regularization parameter value $C$ was found to be more interpretable than one of the other two by humans looking at pre-images form different layers of AlexNet.}
	\label{f:userstudy}
\end{figure}

\subsubsection{Naturalness}\label{s:naturalness}

One of our contributions is the idea that imposing even simple naturalness priors on the reconstructed images improves their interpretability. Since interpretability is a purely subjective attribute, in this section we conduct a small human study to validate this idea. Before that, however, we check whether regularization works as expected and produces images that are statistically closer to natural ones.

\paragraph{Natural image statistics.} Natural images are well known to have certain statistical regularities; for example, the magnitude of image gradients have an exponential distribution~\cite{huang99statistics}. Here we check whether regularized pre-images are more ``natural'' by comparing them to natural images using such statistics. To do so, we compute the histogram of gradient magnitudes for the 100 ImageNet ILSVRC reference images used above and for their AlexNet inversions, for different values of the parameter $C$. Then the original and reconstructed histograms are compared using histogram intersection and their similarity is reported in~\autoref{f:naturalness}. As before, a small amount of regularization is clearly preferable for shallow layers, and a stronger amount is clearly better for intermediate ones. However, the difference is not all that significant for the deepest layers, which are therefore best analyzed in terms of their interpretability.

\paragraph{Interpretability.} In this experiment, pre-images were obtained using the first 25 ILSVRC 2012 validation images as reference. Inversions were obtained from the {mpool1}, {relu3}, {mpool5}, and {fc8} layers of AlexNet for three regularizations strengths: (a) no regularization ($C=\infty$), (b) weak regularization ($C=100$), and (c) strong regularization ($C=1$). Thus we have three pre-images per layer per reference image. In a user study, each subject was shown two randomly picked regularization settings for a layer and reference image. Each subject was asked to select the image that was more interpretable (``whose content is more easily understood''). We conducted this study with 13 human subjects who were not familiar with this line of research, or not familiar with computer vision at all. The first five votes from each subject were discarded to allow them to become familiar with the task. The ordering of images and layers was randomized independently for each subject. Uniform random sampling between the three regularization strengths ensures that no regularization strength dominates the screen or even one side of the screen.

Figure~\ref{f:userstudy} shows the fraction of time a certain regularization strength was found to produce more interpretable results for a given AlexNet layer. Based on these results, at least a small amount of regularization is always preferable for interpretability. Furthermore, strong regularization is highly desirable for very deep layers.

\begin{figure*}
\resizebox{\linewidth}{!}{%
\includelabel{0.124\textwidth}{conv1}{{inversion160_imagenet-caffe-alex_red-fox_layer01-str300}}%
\includelabel{0.124\textwidth}{relu1}{{inversion160_imagenet-caffe-alex_red-fox_layer02-str300}}%
\includelabel{0.124\textwidth}{mpool1}{{inversion160_imagenet-caffe-alex_red-fox_layer03-str300}}%
\includelabel{0.124\textwidth}{norm1}{{inversion160_imagenet-caffe-alex_red-fox_layer04-str300}}%
\includelabel{0.124\textwidth}{conv2}{{inversion160_imagenet-caffe-alex_red-fox_layer05-str300}}%
\includelabel{0.124\textwidth}{relu2}{{inversion160_imagenet-caffe-alex_red-fox_layer06-str300}}%
\includelabel{0.124\textwidth}{mpool2}{{inversion160_imagenet-caffe-alex_red-fox_layer07-str300}}%
\includelabel{0.124\textwidth}{norm2}{{inversion160_imagenet-caffe-alex_red-fox_layer08-str300}}%
}
\resizebox{\linewidth}{!}{%
\includelabel{0.124\textwidth}{conv3}{{inversion160_imagenet-caffe-alex_red-fox_layer09-str300}}%
\includelabel{0.124\textwidth}{relu3}{{inversion160_imagenet-caffe-alex_red-fox_layer10-str300}}%
\includelabel{0.124\textwidth}{conv4}{{inversion160_imagenet-caffe-alex_red-fox_layer11-str100}}%
\includelabel{0.124\textwidth}{relu4}{{inversion160_imagenet-caffe-alex_red-fox_layer12-str100}}%
\includelabel{0.124\textwidth}{conv5}{{inversion160_imagenet-caffe-alex_red-fox_layer13-str20}}%
\includelabel{0.124\textwidth}{relu5}{{inversion160_imagenet-caffe-alex_red-fox_layer14-str20}}%
\includelabel{0.124\textwidth}{fc6}{{inversion160_imagenet-caffe-alex_red-fox_layer16-str01}}%
\includelabel{0.124\textwidth}{relu6}{{inversion160_imagenet-caffe-alex_red-fox_layer17-str01}}%
}
\resizebox{\linewidth}{!}{%
\includelabel{0.124\textwidth}{fc7}{{inversion160_imagenet-caffe-alex_red-fox_layer18-str01}}%
\includelabel{0.124\textwidth}{relu7}{{inversion160_imagenet-caffe-alex_red-fox_layer19-str01}}%
\includelabel{0.124\textwidth}{fc8}{{inversion160_imagenet-caffe-alex_red-fox_layer20-str01}}%
\hspace{0.124\textwidth}%
\hspace{0.124\textwidth}%
\hspace{0.124\textwidth}%
\hspace{0.124\textwidth}%
\hspace{0.124\textwidth}%
}%
\vspace{-0.75em}
\caption{AlexNet inversions (all layers) from the representation of the ``red fox'' image obtained from each layer of AlexNet.}\label{f:inversion-alexnet}
\vspace{0.5em}
\resizebox{\linewidth}{!}{%
\includelabel{0.124\textwidth}{conv1}{{inversion160_imagenet-vgg-m_red-fox_layer01-str300}}%
\includelabel{0.124\textwidth}{conv2}{{inversion160_imagenet-vgg-m_red-fox_layer05-str300}}%
\includelabel{0.124\textwidth}{conv3}{{inversion160_imagenet-vgg-m_red-fox_layer09-str300}}%
\includelabel{0.124\textwidth}{conv4}{{inversion160_imagenet-vgg-m_red-fox_layer11-str100}}%
\includelabel{0.124\textwidth}{conv5}{{inversion160_imagenet-vgg-m_red-fox_layer13-str20}}%
\includelabel{0.124\textwidth}{fc6}{{inversion160_imagenet-vgg-m_red-fox_layer16-str01}}%
\includelabel{0.124\textwidth}{fc7}{{inversion160_imagenet-vgg-m_red-fox_layer18-str01}}%
\includelabel{0.124\textwidth}{fc8}{{inversion160_imagenet-vgg-m_red-fox_layer20-str01}}%
}%
\vspace{-0.75em}
\caption{VGG-M inversions (selected layers). This figure is best viewed in color.}\label{f:inversion-vgg-m}
\vspace{0.5em}
\resizebox{\linewidth}{!}{%
\includelabel{0.124\textwidth}{conv1\_1}{{inversion160_imagenet-vgg-verydeep-16_red-fox_layer01-str300}}%
\includelabel{0.124\textwidth}{conv1\_2}{{inversion160_imagenet-vgg-verydeep-16_red-fox_layer03-str300}}%
\includelabel{0.124\textwidth}{conv2\_1}{{inversion160_imagenet-vgg-verydeep-16_red-fox_layer06-str300}}%
\includelabel{0.124\textwidth}{conv2\_2}{{inversion160_imagenet-vgg-verydeep-16_red-fox_layer08-str300}}%
\includelabel{0.124\textwidth}{conv3\_1}{{inversion160_imagenet-vgg-verydeep-16_red-fox_layer11-str300}}%
\includelabel{0.124\textwidth}{conv3\_2}{{inversion160_imagenet-vgg-verydeep-16_red-fox_layer13-str300}}%
\includelabel{0.124\textwidth}{conv3\_3}{{inversion160_imagenet-vgg-verydeep-16_red-fox_layer15-str300}}%
\includelabel{0.124\textwidth}{conv4\_1}{{inversion160_imagenet-vgg-verydeep-16_red-fox_layer18-str300}}
}
\resizebox{\linewidth}{!}{%
\includelabel{0.124\textwidth}{conv4\_2}{{inversion160_imagenet-vgg-verydeep-16_red-fox_layer20-str300}}%
\includelabel{0.124\textwidth}{conv4\_3}{{inversion160_imagenet-vgg-verydeep-16_red-fox_layer22-str300}}%
\includelabel{0.124\textwidth}{conv5\_1}{{inversion160_imagenet-vgg-verydeep-16_red-fox_layer25-str100}}%
\includelabel{0.124\textwidth}{conv5\_2}{{inversion160_imagenet-vgg-verydeep-16_red-fox_layer27-str20}}%
\includelabel{0.124\textwidth}{conv5\_3}{{inversion160_imagenet-vgg-verydeep-16_red-fox_layer29-str20}}%
\includelabel{0.124\textwidth}{fc6}{{inversion160_imagenet-vgg-verydeep-16_red-fox_layer32-str01}}%
\includelabel{0.124\textwidth}{fc7}{{inversion160_imagenet-vgg-verydeep-16_red-fox_layer34-str01}}%
\includelabel{0.124\textwidth}{fc8}{{inversion160_imagenet-vgg-verydeep-16_red-fox_layer36-str01}}
}%
\vspace{-0.75em}
\caption{VGG-VD-16 inversions (selected layers). This figure is best viewed in color.}\label{f:inversion-vgg-vd}
\end{figure*}

\begin{figure*}[]
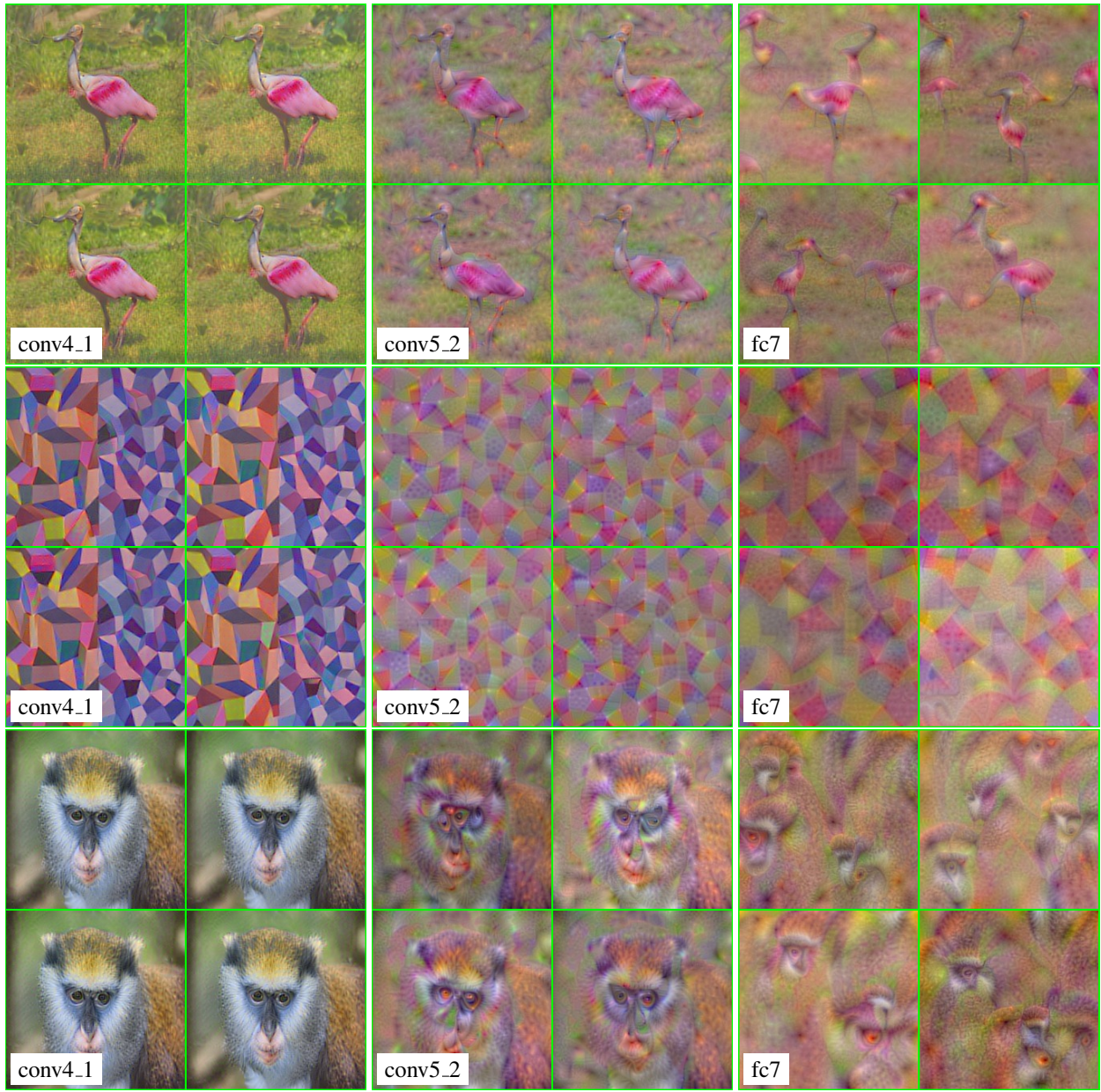

\centering
\resizebox{\linewidth}{!}{%
\includegrid{0.33\textwidth}{2}{conv4\_1}{{inversion-multi-160_imagenet-vgg-verydeep-16_ILSVRC2012_val_00000043_layer20-str300}}\ %
\includegrid{0.33\textwidth}{2}{conv5\_2}{{inversion-multi-160_imagenet-vgg-verydeep-16_ILSVRC2012_val_00000043_layer29-str20}}\ %
\includegrid{0.33\textwidth}{2}{fc7}{{inversion-multi-160_imagenet-vgg-verydeep-16_ILSVRC2012_val_00000043_layer34-str01}}%
}
\resizebox{\linewidth}{!}{%
\includegrid{0.33\textwidth}{2}{conv4\_1}{{inversion-multi-160_imagenet-vgg-verydeep-16_abstract_layer20-str300}}\ %
\includegrid{0.33\textwidth}{2}{conv5\_2}{{inversion-multi-160_imagenet-vgg-verydeep-16_abstract_layer29-str20}}\ %
\includegrid{0.33\textwidth}{2}{fc7}{{inversion-multi-160_imagenet-vgg-verydeep-16_abstract_layer34-str01}}%
}
\resizebox{\linewidth}{!}{%
\includegrid{0.33\textwidth}{2}{conv4\_1}{{inversion-multi-160_imagenet-vgg-verydeep-16_ILSVRC2012_val_00000013_layer20-str300}}\ %
\includegrid{0.33\textwidth}{2}{conv5\_2}{{inversion-multi-160_imagenet-vgg-verydeep-16_ILSVRC2012_val_00000013_layer29-str20}}\ %
\includegrid{0.33\textwidth}{2}{fc7}{{inversion-multi-160_imagenet-vgg-verydeep-16_ILSVRC2012_val_00000013_layer34-str01}}%
}

\caption{For three test images, ``spoonbill'',  ``abstract art'', and ``monkey'', we generate four different reconstructions from layers conv4\_1,  conv5\_2, and fc7 in VGG-VD. This figure is best seen in color.}\label{f:cnn-inversion-multi}
\end{figure*}


\begin{figure*}[ht!]
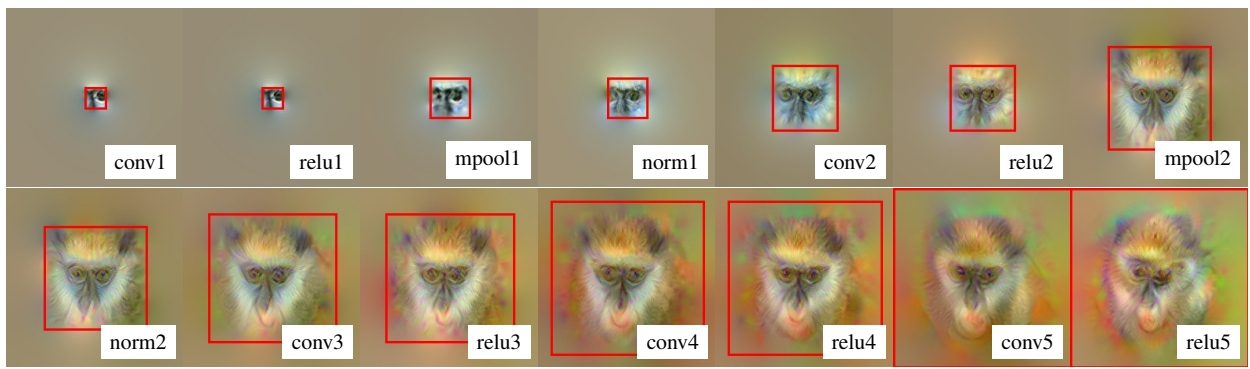

\centering
\newcommand{\putx}[2]{%
\puti{#1}{\noexpand{\adjincludegraphics[width=0.139\textwidth]{inversion-neigh-160-overlaid_imagenet-caffe-ref_ILSVRC2012_val_00000013_layer#2-str01}}}}%
\resizebox{\linewidth}{!}{%
\putx{conv1}{01}%
\putx{relu1}{02}%
\putx{mpool1}{03}%
\putx{norm1}{04}%
\putx{conv2}{05}%
\putx{relu2}{06}%
\putx{mpool2}{07}%
}
\resizebox{\linewidth}{!}{%
\putx{norm2}{08}%
\putx{conv3}{09}%
\putx{relu3}{10}%
\putx{conv4}{11}%
\putx{relu4}{12}%
\putx{conv5}{13}%
\putx{relu5}{14}%
}
\caption{Reconstructions of the ``monkey'' image from a central $5\times 5$ window of feature responses in the convolutional layers of CaffeRef. The red box marks the overall receptive field of the $5\times 5$ window.}\label{f:cnn-neigh}
\end{figure*}
\begin{figure*}[]
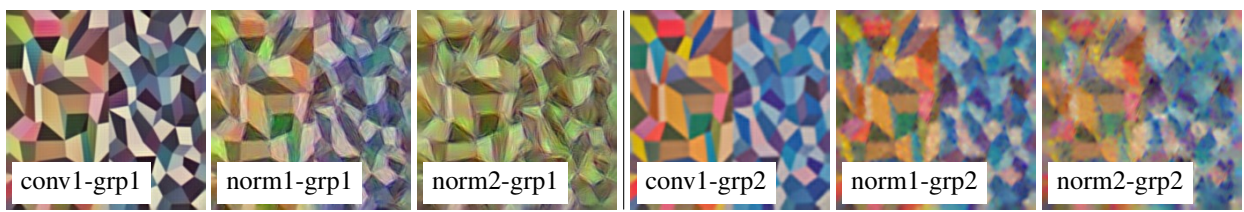

\resizebox{\linewidth}{!}{%
\includelabel{0.16\textwidth}{conv1-grp1}{{inversion-group-160-grp1_imagenet-caffe-ref_abstract_layer01-str100}}\ %
\includelabel{0.16\textwidth}{norm1-grp1}{{inversion-group-160-grp1_imagenet-caffe-ref_abstract_layer04-str20}}\ %
\includelabel{0.16\textwidth}{norm2-grp1}{{inversion-group-160-grp1_imagenet-caffe-ref_abstract_layer08-str20}}%
\ \vrule \ %
\includelabel{0.16\textwidth}{conv1-grp2}{{inversion-group-160-grp2_imagenet-caffe-ref_abstract_layer01-str100}}\ %
\includelabel{0.16\textwidth}{norm1-grp2}{{inversion-group-160-grp2_imagenet-caffe-ref_abstract_layer04-str20}}\ %
\includelabel{0.16\textwidth}{norm2-grp2}{{inversion-group-160-grp2_imagenet-caffe-ref_abstract_layer08-str20}}\ %
}
\caption{{\bf CNN neural streams.} Reconstructions of the ``abstract'' test image from either of the two neural streams in CaffeRef. This figure is best seen in color.}\label{f:cnn-streams}
\end{figure*}

\subsubsection{Inversion of different layers}

Having established the legitimacy of the inversions, next we study qualitatively the reconstructions obtained from different layers of the three CNNs for a test image (``red fox''). In particular,  \autoref{f:inversion-alexnet} shows the reconstructions obtained from each layer of AlexNet and \autoref{f:inversion-vgg-m} and~\autoref{f:inversion-vgg-vd} do the same for all the linear (convolutional and fully connected) layers of VGG-M and VGG-VD.

The progression is remarkable. The first few layers of all the networks compute a code of the image that is nearly exactly invertible. All the layers prior to the fully-connected ones preserve instance-specific details of the image, although with increasing fuzziness. The 4,096-dimensional fully connected layers discard more geometric as well as instance-specific information as they invert back to a \emph{composition of parts, which are similar but not identical to the ones found in the original image}. Unexpectedly, even the very last layer, fc8, whose 1,000 components are in principle category predictors, still appears to preserve some instance-specific details of the image.

Comparing different architectures, VGG-M reconstructions are sharper and more detailed than the ones obtained from AlexNet, as it may be expected due to the denser and higher dimensional filters used here (compare for example conv4 in \autoref{f:inversion-alexnet} and \autoref{f:inversion-vgg-m}).  VGG-VD emphasizes these differences more. First, abstractions are achieved much more gradually in this architecture: conv5\_1, conv5\_2 and conv5\_3 reconstructions resemble the reconstructions from conv5 in AlexNet and VGG-M, despite the fact that there are three times more intermediate layers in VGG-VD. Nevertheless, fine details are preserved more accurately in the deep layers in this architecture (compare for example, the nose and eyes of the fox in conv5 in VGG-M and conv5\_1 -- conv5\_3 in VGG-VD).

Another difference we noted here, as well as in \autoref{f:category}, is that reconstructions from deep VGG-VD layers are often more zoomed in compared to other networks (see for example the ``abstract art'' and ``monkey'' reconstructions from fc7 in \autoref{f:cnn-inversion-multi} and, for activation maximization, in \autoref{f:category}). The preference of VGG-VD for large, detailed object occurrences may be explained by its better ability to represent fine-grained object details, such as textures.

\subsubsection{Reconstruction ambiguity and invariances}

\autoref{f:cnn-inversion-multi} examines the invariances captured by the VGG-VD codes by comparing multiple reconstructions obtained from several deep layers.
A careful examination of these images reveals that the codes capture progressively larger deformations of the object.
In the ``spoonbill'' image, for example, conv5\_2 reconstructions show slightly different body poses, evident from the different leg configurations. In the ``abstract art'' test image, a close examination of the pre-images reveals that, while the texture statistics are preserved well, the instance-specific details are in fact completely different in each image: the location of the vertexes, the number and orientation of the edges, and the color of the patches are not the same at the same locations in different images. This case is also remarkable as the training data for VGG-VD, i.e. ImageNet ILSVRC, does not contain any such pattern suggesting that these codes are indeed rather generic. Inversions from fc7 result in multiple copies of the object/parts at different positions and scales for the ``spoonbill'' and ``monkey'' cases. For the ``monkey'' and ``abstract art'' cases, inversions from fc7 appear to result in slightly magnified versions of the pattern: for instance, the reconstructed monkey's eye is about 20\% larger than in the original image; and the reconstructed patches in ``abstract art'' are about 70\% larger than in the original image. The preference for reconstructing larger object scales seems to be typical of VGG-VD (see also \autoref{f:category}).

Note that all these reconstructions and the original images are very similar from the viewpoint of the CNN representation; we conclude in particular that the deepest layers find the original images and a number of scrambled parts to be equivalent. This may be considered as another type of natural confounder for CNNs alternative to those discussed in~\cite{nguyen15deep}.

\subsubsection{Reconstruction biases}

It is interesting to note that some of the inverted images have large green regions (for example see \autoref{f:inversion-vgg-vd} fc6 to fc8).
This property is likely to be intrinsic to the networks and not induced, for example, by the choice of natural image prior, the effect of which is demonstrated in~\autoref{fig:smoothing} and \autoref{s:jitter}.
The prior only encourages smoothness as it is equivalent (for $\beta=2$) to penalizing high-frequency components of the reconstructed image.
Importantly, the prior is applied to all color channels equally.
When gradually removing the prior, random high-frequency components dominate and it is harder to discern a human-interpretable signal.

\subsubsection{Inversion from selected representation components}

It is also possible to examine reconstructions obtained from subsets of neural responses in different CNN layers.
\autoref{f:cnn-neigh} explores the \emph{locality} of the codes by reconstructing a central $5\times 5$ patch of features in each layer.
The regularizer encourages portions of the image that do not contribute to the neural responses to be switched off.
The locality of the features is obvious in the figure; what is less obvious is that the effective receptive field of the neurons is in some cases significantly smaller than the theoretical one shown as a red box in the image.

Finally, \autoref{f:cnn-streams} reconstructs images from two different subsets of feature channels for CaffeRef. These subsets are induced by the fact that the first several layers (up to norm2) of the CaffeRef architecture are trained to have blocks of independent filters~\cite{krizhevsky12imagenet}. Reconstructing individually from each subset clearly shows that one group is tuned towards color information, whereas the second one is tuned towards sharper edges and luminance components. Remarkably, this behavior emerges spontaneously in the learned network.

\section{Visualization by activation maximization}\label{s:results-activation-maximization}

In this section the activation maximization method is applied to classical and CNN representations.

\begin{figure*}[t]
\centering%
\resizebox{\textwidth}{!}{%
	\includegraphics[height=0.18\textwidth]{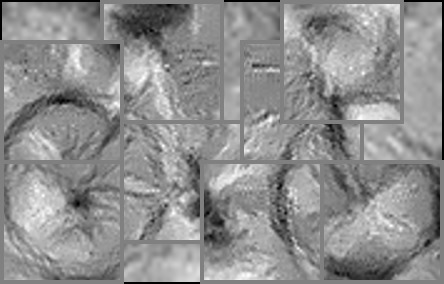}\ \ %
	\includegraphics[height=0.18\textwidth]{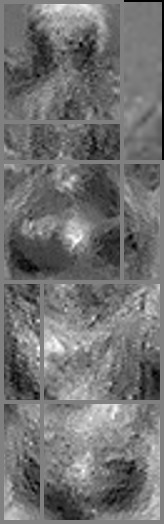}\ \ %
	\includegraphics[height=0.18\textwidth]{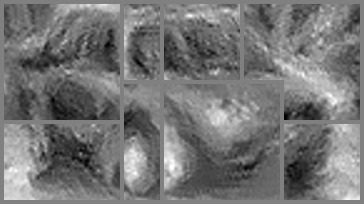}\ \ %
	\includegraphics[height=0.18\textwidth]{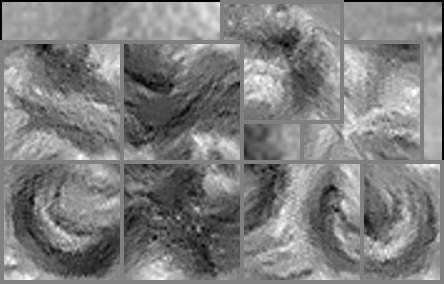}\ \ %
	\includegraphics[height=0.18\textwidth]{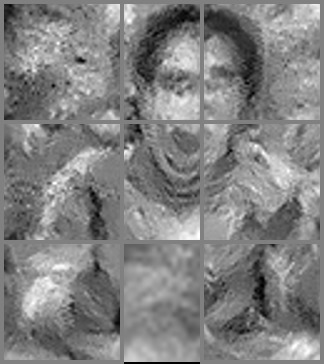}\ \ %
	\includegraphics[height=0.18\textwidth]{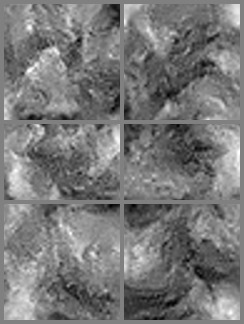}
} 	
\\[0.3em]

\resizebox{\textwidth}{!}{%
\includegraphics[height=0.18\textwidth]{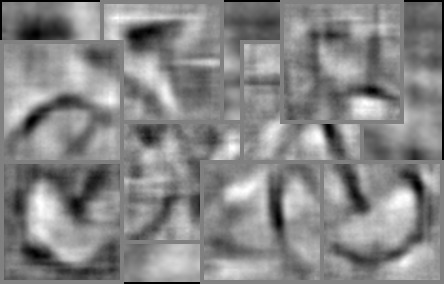}\ \ %
\includegraphics[height=0.18\textwidth]{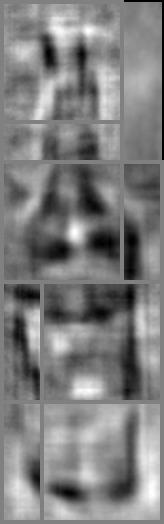}\ \ %
\includegraphics[height=0.18\textwidth]{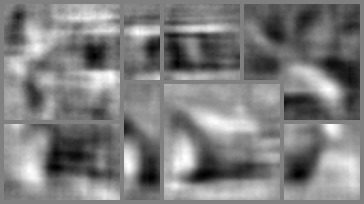}\ \ %
\includegraphics[height=0.18\textwidth]{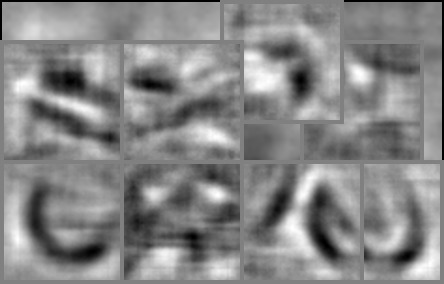}\ \ %
\includegraphics[height=0.18\textwidth]{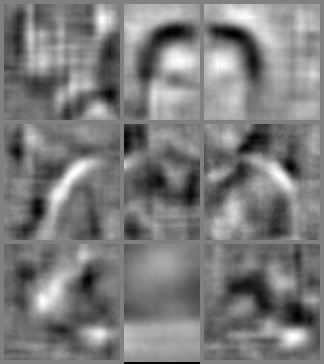}\ \ %
\includegraphics[height=0.18\textwidth]{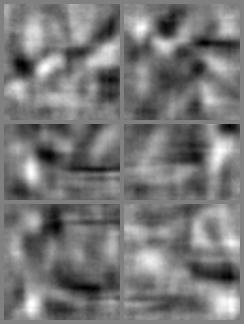}
}

\caption{Visualization of DPMv5 HOG models using activation maximization (top) and HOGgle (bottom). Each model comprises a ``root'' filter overlaid with several part filters. From left to right: Bicycle (Component 2), Bottle (5), Car (4), Motorbike (2), Person (1), Potted Plant (4).}\label{f:hogmodels}

\end{figure*}

\subsection{Classical representations}

For classical representations, we use activation maximization to visualize HOG templates. Let $\Phi_\mathrm{HOG}(\bx)$ denote the HOG descriptor of a gray scale image $\bx \in \real^{H\times W}$; a \emph{HOG template} is a vector $\bw$, usually learned by a latent SVM, that defines a scoring function $\Phi(\bx) = \langle \bw, \Phi_\mathrm{HOG}(\bx) \rangle$ for a particular object category. The function $\Phi(\bx)$ can be interpreted as a CNN consisting of the HOG CNN $\Phi_\mathrm{HOG}(\bx)$ followed by a linear projection layer of parameter $\bw$. The output of $\Phi(\bx)$ is a scalar, expressing the confidence that the image $\bx$ contains the target object class.

In order to visualize the template $\bw$ using activation maximization, the loss function $- \langle \Phi(\bx), \Phi_0 \rangle/Z$ of~\autoref{e:objective3} is plugged in the objective function~\eqref{e:objective}. Since in this case $\Phi(\bx)$ is a scalar function, the reference vector $\Phi_0$ is also a scalar, which is set to 1. The normalization constant $Z$ is set to
\begin{equation}\label{e:zactivationmaximization}
Z = M \rho^2
\end{equation}
where $\rho = \max\{H,W\}$ and $M$ is an estimate of range of $\Phi(\bx)$, obtained as
$
M = \langle |\bw|, \Phi_\mathrm{HOG}(\bx) \rangle
$
where $|\bw|$ is the element wise absolute value of $\bw$ and $\bx$ is set to a white noise sample. The method is used to visualize the DPM (v5) models~\cite{voc-release5} trained on the VOC2010~\cite{pascal-voc-2010} dataset. 

These visualizations are compared to the ones obtained in the analogous experiment by Vondrick~\etal~(\cite{vondrick13hoggles:}~Fig.~14) using HOGgle. An important difference is that HOGgle does not perform activation maximization, but rather inversion and returns an approximate pre-image $\Phi^{-1}_\text{HOG}(\bw_+)$, where $\bw_+ = \max\{0,\bw\}$ is the rectified template. Strictly speaking, inversion is not applicable here because the template $\bw$ is \emph{not} a HOG descriptor. In particular, $\bw$ contains negative components which HOG descriptors do not contain. Even after rectification, there is usually no image such that $\Phi_\text{HOG}(\bx) = \bw_+$. By contrast, activation maximization is principled because it works on top of the detector scoring function; in this manner it can correctly reflect the effect of both positive and negative components in $\bw$.

The visualizations of five DPMs using activation maximization and HOGgle are shown in \autoref{f:hogmodels}. Note that the DPMs are hierarchical, and consist of a root object template and several higher-resolution part templates. For simplicity, each part is processed independently, but activation maximization could be applied to the composition of all the parts to remove the seams between them.

Compared to HOGgle, activation maximization reconstructs finer details, as can be noted for parts such as the bicycle wheel and bottle top. On the other hand, HOGgle reconstructions contain stronger and straighter edges (see for example the car roof). The latter may be a result of HOGgle using a restricted dictionary computed from natural images whereas our approach uses a more generic smoothness prior.

\begin{figure*}[]
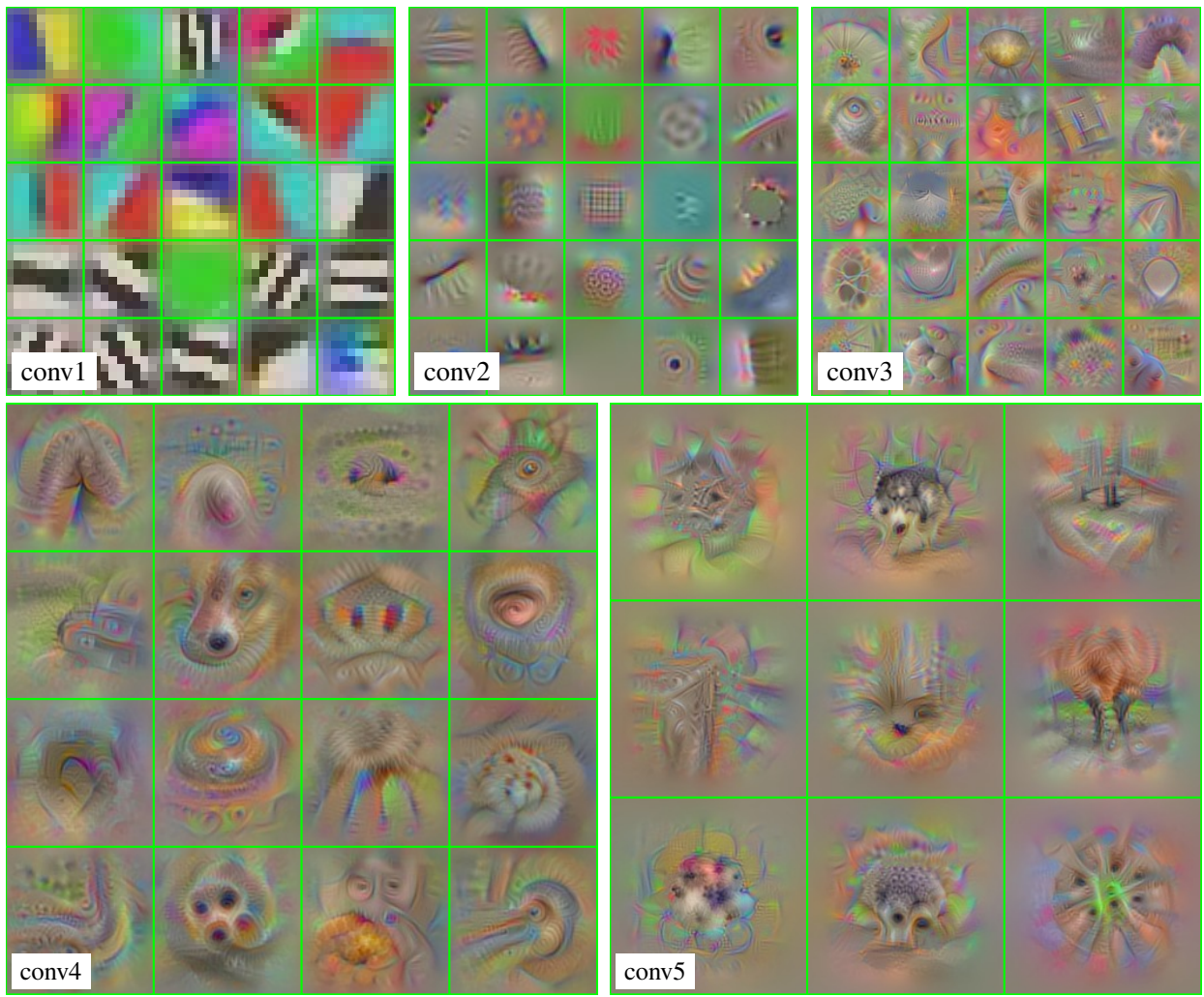
\centering%
%
\resizebox{\textwidth}{!}{%
\includegrid[3pt 4pt 4pt 3pt]{0.3\textwidth}{5}{conv1}{view160_imagenet-vgg-m-l01-mosaic}\ \ %
\includegrid[7pt 11pt 11pt 7pt]{0.3\textwidth}{5}{conv2}{view160_imagenet-vgg-m-l05-mosaic}\ \ %
\includegrid[7pt 11pt 11pt 7pt]{0.3\textwidth}{5}{conv3}{view160_imagenet-vgg-m-l09-mosaic}
}
\\[0.2em]
\resizebox{\textwidth}{!}{%
\includegrid[12pt 126pt 126pt 12pt]{0.49\textwidth}{4}{conv4}{view160_imagenet-vgg-m-l11-mosaic}\ \ %
\includegrid[12pt 370pt 370pt 12pt]{0.49\textwidth}{3}{conv5}{view160_imagenet-vgg-m-l13-mosaic}
}
\vspace{-1em}
\caption{Activation maximization of the first filters of each convolutional layer in VGG-M.}\label{f:vgg-m-conv}
\end{figure*}
\begin{figure*}[]
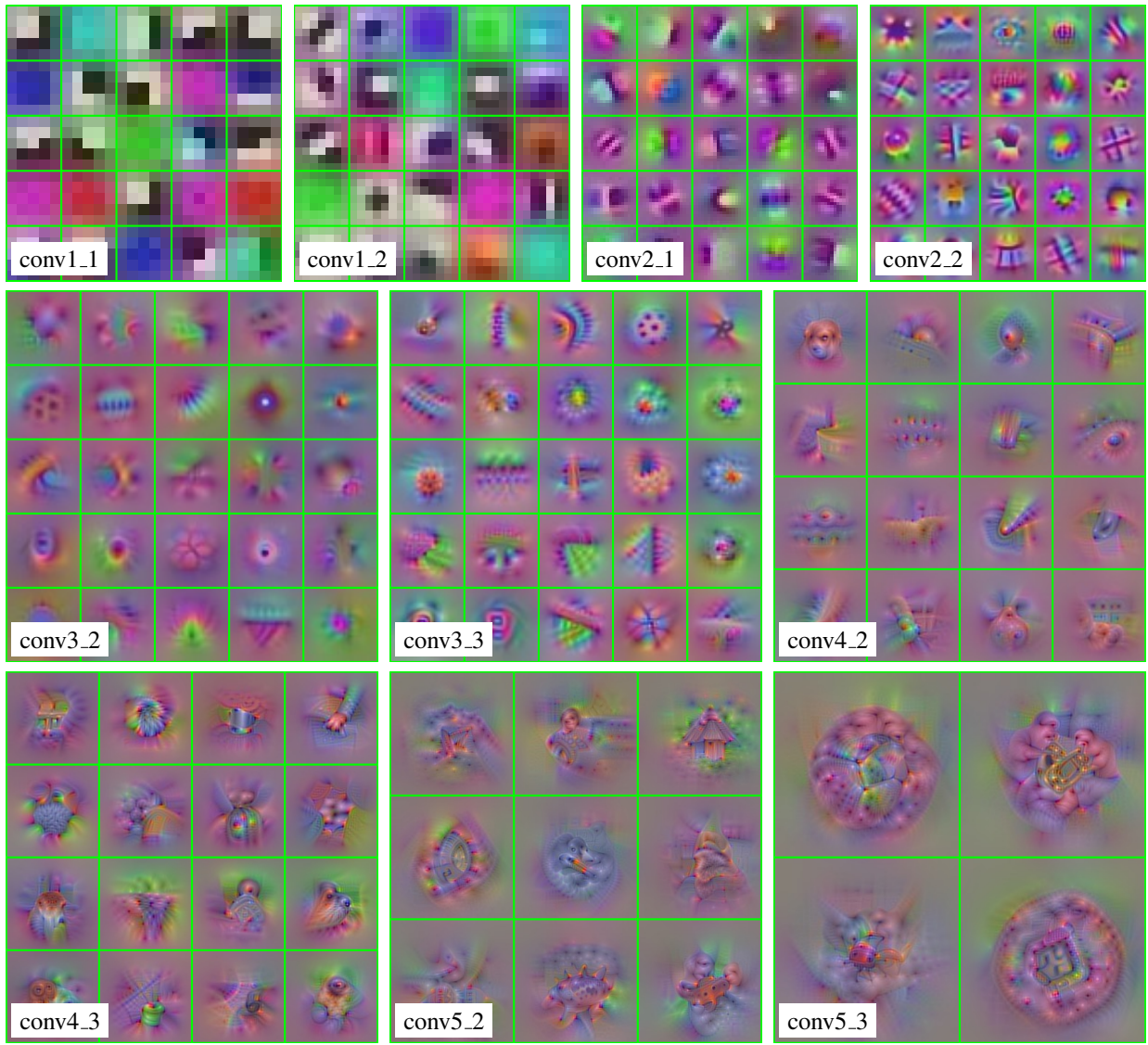
\centering%
\resizebox{\linewidth}{!}{
\includegrid{0.24\textwidth}{5}{conv1\_1}{view160_imagenet-vgg-verydeep-16-l01-mosaic}\ \ %
\includegrid{0.24\textwidth}{5}{conv1\_2}{view160_imagenet-vgg-verydeep-16-l03-mosaic}\ \ %
\includegrid{0.24\textwidth}{5}{conv2\_1}{view160_imagenet-vgg-verydeep-16-l08-mosaic}\ \ %
\includegrid{0.24\textwidth}{5}{conv2\_2}{view160_imagenet-vgg-verydeep-16-l11-mosaic}
}
\\[0.3em]

\resizebox{\linewidth}{!}{
\includegrid{0.32\textwidth}{5}{conv3\_2}{view160_imagenet-vgg-verydeep-16-l13-mosaic}\ \ %
\includegrid{0.32\textwidth}{5}{conv3\_3}{view160_imagenet-vgg-verydeep-16-l15-mosaic}\ \ %
\includegrid[0pt 95pt 95pt 0pt]{0.32\textwidth}{4}{conv4\_2}{view160_imagenet-vgg-verydeep-16-l20-mosaic}
}
\\[0.3em]

\resizebox{\linewidth}{!}{

\includegrid[0pt 110pt 110pt 0pt]{0.32\textwidth}{4}{conv4\_3}{view160_imagenet-vgg-verydeep-16-l22-mosaic}\ \ %
\includegrid[0pt 370pt 370pt 0pt]{0.32\textwidth}{3}{conv5\_2}{view160_imagenet-vgg-verydeep-16-l27-mosaic}\ \ %
\includegrid[0pt 640pt 640pt 0pt]{0.32\textwidth}{2}{conv5\_3}{view160_imagenet-vgg-verydeep-16-l29-mosaic}
}

\caption{Activation maximization of the first filters for each convolutional layer in VGG-VD-16.}\label{f:vgg-verydeep-16-conv}
\end{figure*}
\begin{figure*}[]
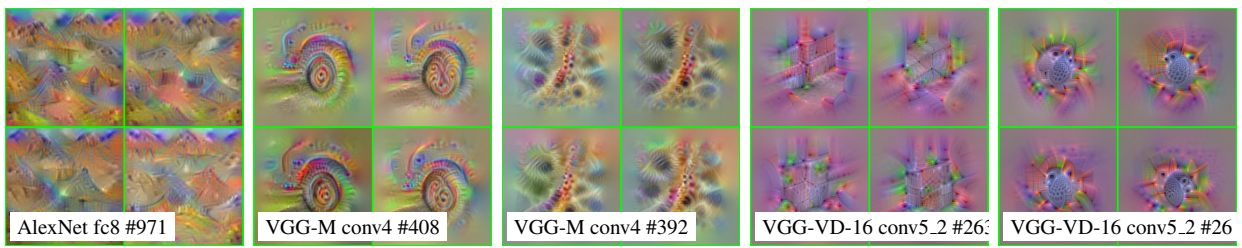
\centering%
\resizebox{\linewidth}{!}{
\includegrid{0.24\textwidth}{2}{AlexNet fc8 \#971}{view160-noleak-multi_imagenet-caffe-alex-l20_f0971}\ \ %
\includegrid{0.24\textwidth}{2}{VGG-M conv4 \#408}{view160-noleak-multi_imagenet-vgg-m-l11_f0408}\ \ %
\includegrid{0.24\textwidth}{2}{VGG-M conv4 \#392}{view160-noleak-multi_imagenet-vgg-m-l11_f0392}\ \ %
\includegrid{0.24\textwidth}{2}{VGG-VD-16 conv5\_2 \#263}{view160-noleak-multi_imagenet-vgg-verydeep-16-l27_f0263}\ \ %
\includegrid{0.24\textwidth}{2}{VGG-VD-16 conv5\_2 \#26}{view160-noleak-multi_imagenet-vgg-verydeep-16-l27_f0026}
}
\caption{Activation maximization with 4 different initializations.}\label{f:multimax}
\end{figure*}
\begin{figure*}[]
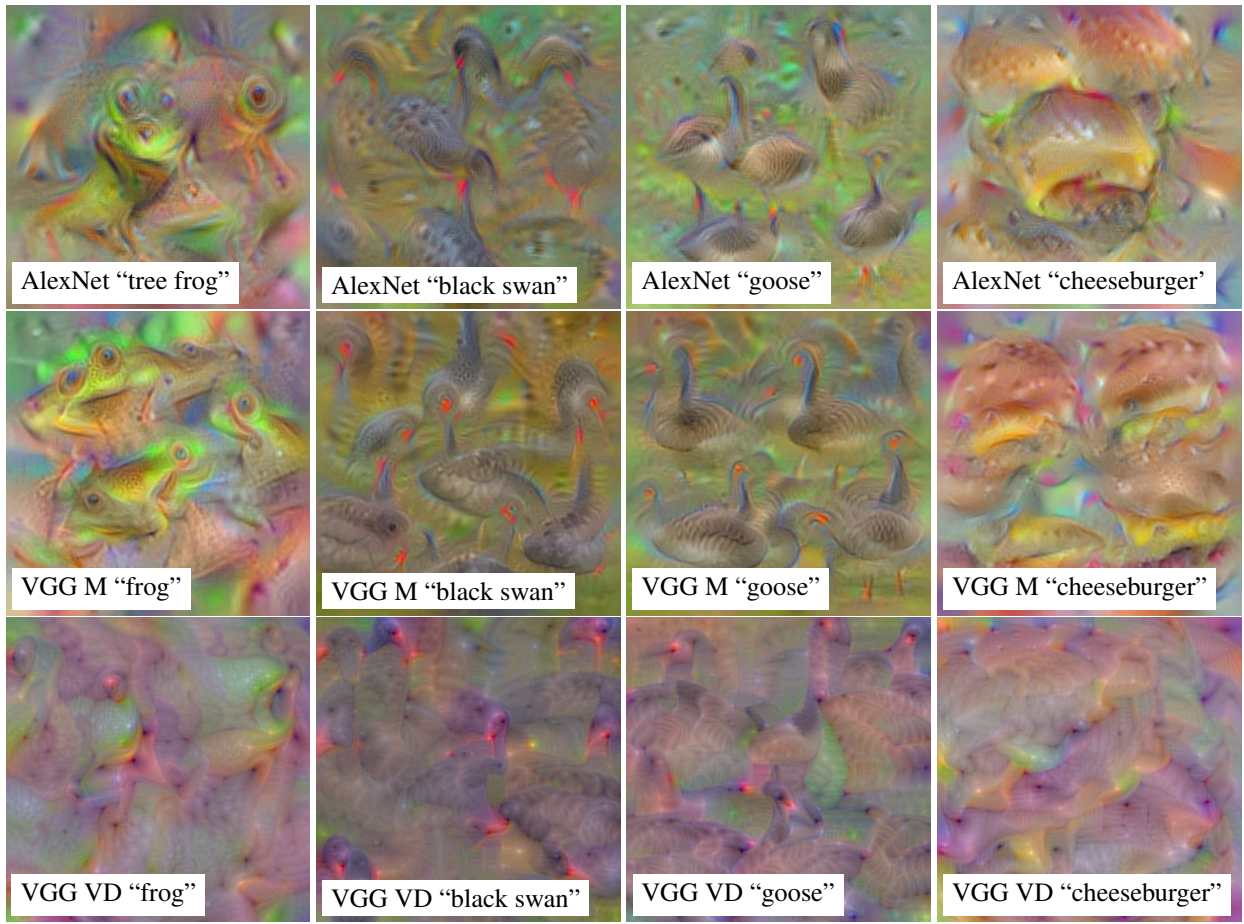

\resizebox{\textwidth}{!}{%
\includelabel{0.245\textwidth}{AlexNet ``tree frog''}{category160_imagenet-caffe-alex_treefrog}\ %
\includelabel{0.245\textwidth}{AlexNet ``black swan''}{category160_imagenet-caffe-alex_blackswan}\ %
\includelabel{0.245\textwidth}{AlexNet ``goose''}{category160_imagenet-caffe-alex_goose}\ %
\includelabel{0.245\textwidth}{AlexNet ``cheeseburger'}{category160_imagenet-caffe-alex_cheeseburger}
}
\resizebox{\textwidth}{!}{%
\includelabel{0.245\textwidth}{VGG M ``frog''}{category160_imagenet-vgg-m_treefrog}\ %
\includelabel{0.245\textwidth}{VGG M ``black swan''}{category160_imagenet-vgg-m_blackswan}\ %
\includelabel{0.245\textwidth}{VGG M ``goose''}{category160_imagenet-vgg-m_goose}\ %
\includelabel{0.245\textwidth}{VGG M ``cheeseburger''}{category160_imagenet-vgg-m_cheeseburger}
}
\resizebox{\textwidth}{!}{%
\includelabel{0.245\textwidth}{VGG VD ``frog''}{category160_imagenet-vgg-verydeep-16_treefrog}\ %
\includelabel{0.245\textwidth}{VGG VD ``black swan''}{category160_imagenet-vgg-verydeep-16_blackswan}\ %
\includelabel{0.245\textwidth}{VGG VD ``goose''}{category160_imagenet-vgg-verydeep-16_goose}\ %
\includelabel{0.245\textwidth}{VGG VD ``cheeseburger''}{category160_imagenet-vgg-verydeep-16_cheeseburger}
}
\caption{Activation maximization for the second to last layer of AlexNet, VGG-M, VGG-VD-16 for the classes ``frog'', ``black swan'', ``goose'', and ``vending machine''. The second to last layer codes directly for different classes, before softmax normalization.}\label{f:category}
\end{figure*}

\subsection{CNN representations}

Next, the activation maximization method is applied to the study of deep CNNs. As before, the inner-product loss~\autoref{e:objective3} is used in the objective~\eqref{e:objective}, but this time the reference vector $\Phi_0$ is set to the one-hot indicator vector of the representation component being visualized. It was not possible to find an architecture independent normalization constant $Z$ in~\autoref{e:objective3} as different CNNs have very different ranges of neuron output. Instead Z is calculated in the same way as \autoref{e:zactivationmaximization} where $M$ is the maximum value achieved by the representation component in the ImageNet ILSVRC 2012 validation data and $\rho$ the size of the component receptive field, as reported in \autoref{f:allnetworks}.
As in \autoref{s:result-inversion-cnn} the jitter amount $T$ in (\autoref{s:jitter}) is set to a fourth of the stride of the feature and all the other parameters are set as described in \autoref{s:balancing} (including $C=1$).

\autoref{f:vgg-m-conv} shows the visual patterns obtained by maximally activating the few components in the convolutional layers of VGG-M. Similarly to~\cite{zeiler14visualizing} and~\cite{yosinksi15understanding}, the complexity of the patterns increases substantially with depth. The first convolutional layer conv1 captures colored edges and blobs, but the complexity of the patterns generated by conv3, conv4 and conv5 is remarkable. While some of these pattern do evoke objects or object parts, it remains difficult to associate to them a clear semantic interpretation (differently from~\cite{yosinksi15understanding} we prefer to avoid hand-picking semantically interpretable filters). This is not entirely surprising given that the representation is distributed and activations may need to be combined to form a meaning. Experiments from AlexNet yielded entirely analogous if a little blurrier results.

\autoref{f:vgg-verydeep-16-conv} shows the patterns obtained from VGG-VD. The complexity of the patterns build up more gradually than for VGG-M and AlexNet. Qualitatively, the complexity of the stimuli in conv5 in AlexNet and VGG-M seems to be comparable to conv4\_3 and conv5\_1 in VGG-VD. conv5\_2 and conv5\_3 do appear to be significantly more complex, however.  A second observation is that the reconstructed colors tend to be much more saturated, probably due to the lack of normalization layers in the architecture. Thirdly, we note that reconstructions contain significantly more fine-grained details, and in particular tiny blob-like structures, which probably activate very strongly the first very small filters in the network.

\autoref{f:category} repeats the experiment from~\cite{simonyan14deep}, more recently reprised by~\cite{yosinksi15understanding} and~\cite{mordvintsev15inceptionism:}, and maximizes the component of fc8 that correspond to a given class prediction. Four classes are considered: two similar animals (``black swan'' and ``goose''), a different one (``tree frog''), and an inanimate object (``cheeseburger''). We note that in all cases it is easy to identify several parts or even instances of the target object class. However, reconstructions are fragmented and scrambled, indicating that the representations are highly invariant to occlusions and pose changes. Secondly, reconstructions from VGG-M are considerably sharper than the ones obtained from AlexNet, as could be expected. Thirdly, VGG-VD-16 differs significantly from the other two architectures. Colors are more ``washed out'', which we impute to the lack of normalization in the architecture as for~\autoref{f:vgg-verydeep-16-conv}. Reconstructions tends to focus on much larger objects; for example, the network clearly captures the feather pattern of the bird as well as the rough skin of the frog.

Finally, \autoref{f:multimax} shows multiple pre-images of a few representation components obtained by starting activation maximization from different random initializations. This is analogous to \autoref{f:cnn-inversion-multi} and is meant to probe the invariances in the representation. The variation across the four pre-images is a mix of geometric and style transformations. For example, the second neuron appears to represent four different variants of a wheel of a vehicle.

\begin{figure*}[]
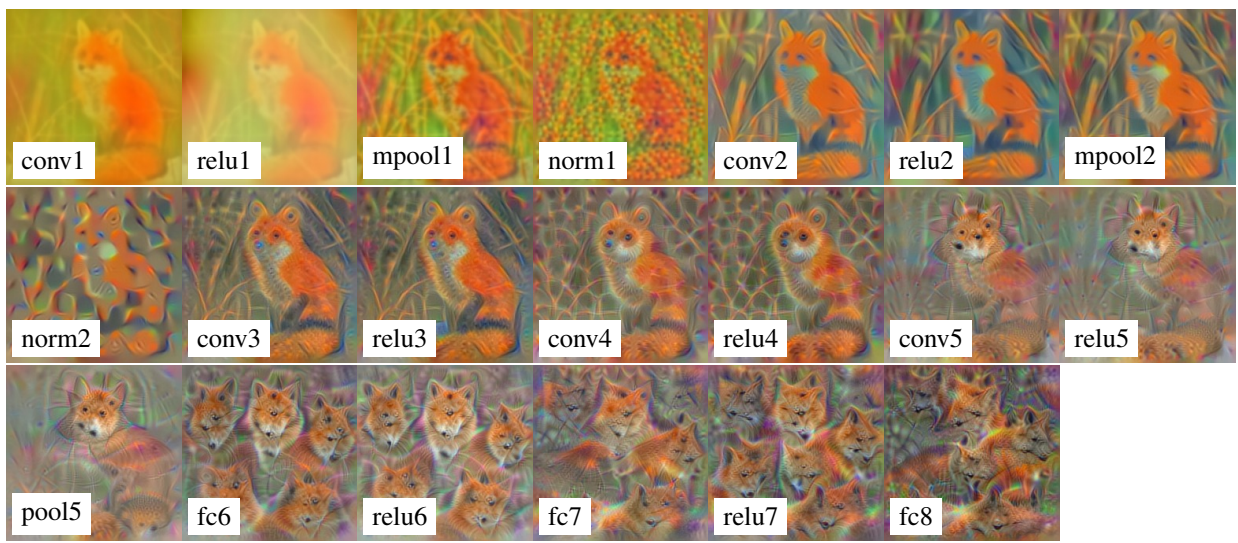

\resizebox{\linewidth}{!}{
\includelabel{0.14\textwidth}{conv1}{enhance160_imagenet-vgg-m_layer1_red-fox}%
\includelabel{0.14\textwidth}{relu1}{enhance160_imagenet-vgg-m_layer2_red-fox}%
\includelabel{0.14\textwidth}{mpool1}{enhance160_imagenet-vgg-m_layer3_red-fox}%
\includelabel{0.14\textwidth}{norm1}{enhance160_imagenet-vgg-m_layer4_red-fox}%
\includelabel{0.14\textwidth}{conv2}{enhance160_imagenet-vgg-m_layer5_red-fox}%
\includelabel{0.14\textwidth}{relu2}{enhance160_imagenet-vgg-m_layer6_red-fox}%
\includelabel{0.14\textwidth}{mpool2}{enhance160_imagenet-vgg-m_layer7_red-fox}
}
\resizebox{\linewidth}{!}{
\includelabel{0.14\textwidth}{norm2}{enhance160_imagenet-vgg-m_layer8_red-fox}%
\includelabel{0.14\textwidth}{conv3}{enhance160_imagenet-vgg-m_layer9_red-fox}%
\includelabel{0.14\textwidth}{relu3}{enhance160_imagenet-vgg-m_layer10_red-fox}%
\includelabel{0.14\textwidth}{conv4}{enhance160_imagenet-vgg-m_layer11_red-fox}%
\includelabel{0.14\textwidth}{relu4}{enhance160_imagenet-vgg-m_layer12_red-fox}%
\includelabel{0.14\textwidth}{conv5}{enhance160_imagenet-vgg-m_layer13_red-fox}%
\includelabel{0.14\textwidth}{relu5}{enhance160_imagenet-vgg-m_layer14_red-fox}
}
\resizebox{\linewidth}{!}{
\includelabel{0.14\textwidth}{pool5}{enhance160_imagenet-vgg-m_layer15_red-fox}%
\includelabel{0.14\textwidth}{fc6}{enhance160_imagenet-vgg-m_layer16_red-fox}%
\includelabel{0.14\textwidth}{relu6}{enhance160_imagenet-vgg-m_layer17_red-fox}%
\includelabel{0.14\textwidth}{fc7}{enhance160_imagenet-vgg-m_layer18_red-fox}%
\includelabel{0.14\textwidth}{relu7}{enhance160_imagenet-vgg-m_layer19_red-fox}%
\includelabel{0.14\textwidth}{fc8}{enhance160_imagenet-vgg-m_layer20_red-fox}%
\hspace{0.14\textwidth}
}
\caption{Caricatures of the ``red fox'' image obtained from the different layers in VGG-M.}\label{f:caricature-vgg-m}
\end{figure*}

\begin{figure*}[]
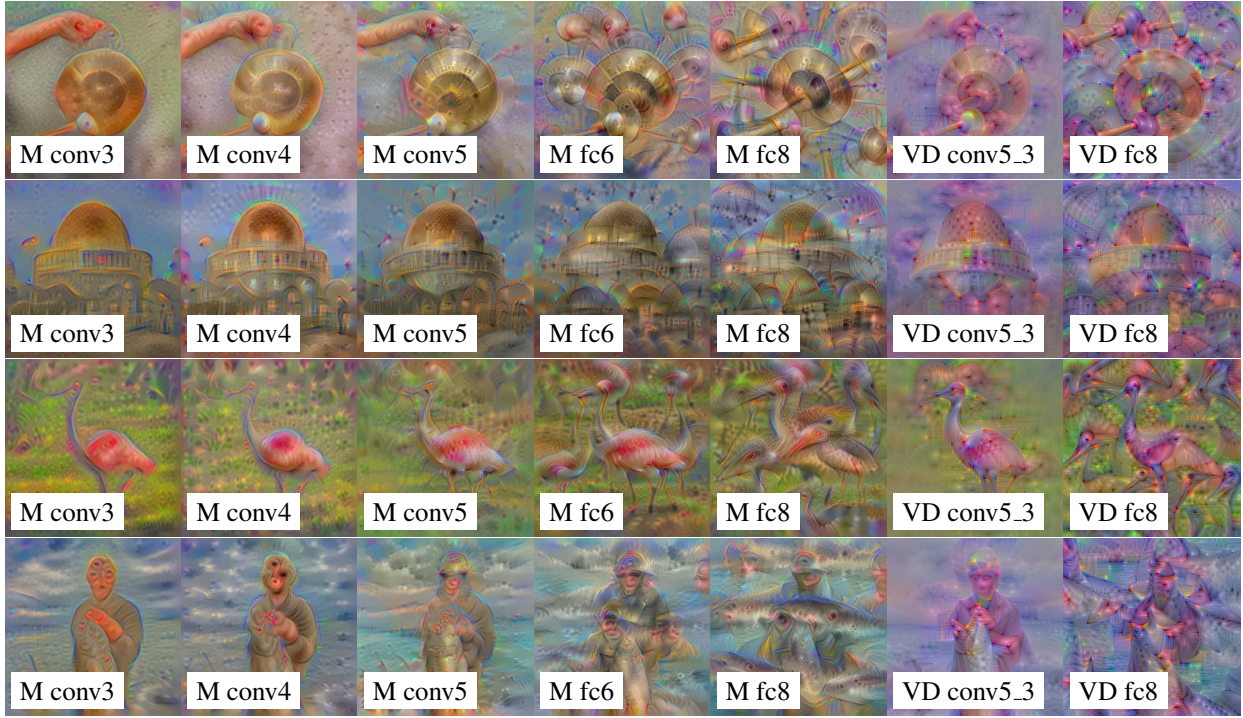

\newcommand{\doone}[1]{%
\resizebox{\linewidth}{!}{%
\includelabel{0.14\textwidth}{M conv3}{enhance160_imagenet-vgg-m_layer9_#1}%
\includelabel{0.14\textwidth}{M conv4}{enhance160_imagenet-vgg-m_layer11_#1}%
\includelabel{0.14\textwidth}{M conv5}{enhance160_imagenet-vgg-m_layer13_#1}%
\includelabel{0.14\textwidth}{M fc6}{enhance160_imagenet-vgg-m_layer16_#1}%
\includelabel{0.14\textwidth}{M fc8}{enhance160_imagenet-vgg-m_layer20_#1}%
%
\includelabel{0.14\textwidth}{VD conv5\_3}{enhance160_imagenet-vgg-verydeep-16_layer29_#1}%
\includelabel{0.14\textwidth}{VD fc8}{enhance160_imagenet-vgg-verydeep-16_layer36_#1}
}
}
\doone{gong}
\doone{other_ILSVRC2012_val_00000024}
\doone{other_ILSVRC2012_val_00000043}
\doone{fish}
\caption{Caricatures of a number of test images obtained from the different layers of VGG-M (conv3, conv4, conv5, fc6, fc8) and VGG-VD (conv5\_3 and fc8).}\label{f:caricatures}
\end{figure*}

\section{Visualization by caricaturization}\label{s:results-activation-highlighting}

Our last visualization is inspired by Google's Inceptionism~\cite{mordvintsev15inceptionism:}. It is similar to activation maximization (\autoref{s:results-activation-maximization}) and in fact uses the same formulation for~\autoref{e:objective} with the inner-product loss~\autoref{e:objective3}. However,  there are two key differences. First, the target mask is now set to
\[
 \Phi_0 = \max\{0, \Phi(\bx_0) \}
\]
where $\bx_0$ is a reference image and the normalization factor $Z$ is set to $\|\Phi_0\|^2$. Second, the optimization is started from the image $\bx_0$ itself.

The idea of this visualization is to exaggerate any pattern in $\bx_0$ that is active in the representation $\Phi(\bx_0)$, hence creating a ``caricature'' of the image according to this model.  Furthermore, differently from activation maximization, this visualization works with combinations of multiple activations instead of individual ones.

\autoref{f:caricature-vgg-m} shows the caricatures of the ``red-fox'' image obtained from the different layers of VGG-M. Applied to the first block of layers, the procedure simply saturates the color. conv2 appears to be tuned to long, linear structures, conv4 to round ones, and conv5 to the head (part) of the fox. The fully connected layers generate mixtures of fox heads, including hallucinating several in the background, as already noted in~\cite{mordvintsev15inceptionism:}. \autoref{f:caricatures} shows the caricatures obtained from selected layers of VGG-M and VGG-VD, with similar results.
\section{Summary}\label{s:summary}

There is a growing interest in methods that can help us understand computer vision representations, and in particular representations learned automatically from data such as those constructed by deep CNNs. Recently, several authors have proposed complementary visualization techniques to do so. In this manuscript we have extended our previous work on inverting representations using natural pre-images to a unified framework that encompasses several visualization types.  We have then experimented with three such visualizations (inversion, activation maximization, and caricaturization), and used those to probe and compare standard classical representations, and CNNs.

The robustness of our visualization method has been assessed quantitatively in the case of the inversion problem by comparing the output of  our approach to earlier feature inversion techniques. The most important results, however, emerged from an analysis of the visualizations obtained from deep CNNs; some of these are: the fact that photometrically accurate information is preserved deep down in CNNs, that even very deep layers contain instance-specific information about objects, that intermediate convolutional layers capture local invariances to pose and fully connected layers to large variations in the object layouts, that individual CNN components  code for complex but, for the most part, not semantically-obvious patterns, and that different CNN layers appear to capture different types of structures in images, from lines and curves to parts.

We believe that these visualization methods can be used as direct diagnostic tools to further research in CNNs. For example, an interesting problem is to look for semantically-meaningful activation patterns in deep CNN layers (given that individual responses are often not semantic); inversion, or variants of activation maximization, can be used to validate such activation patterns by means of visualizations.


\section*{Acknowledgements}

We gratefully acknowledge the support of the ERC StG IDIU for Andrea Vedaldi and of BP for Aravindh Mahendran.

\bibliographystyle{spmpsci}
\bibliography{local}
\end{document}